\documentclass[10pt,journal,compsoc,twoside]{IEEEtran}
\usepackage{amsmath,amssymb,amsthm}
\usepackage{graphicx}
\usepackage{color}
\usepackage{times}
\usepackage{subfig}
\usepackage{url}

\input{Definitions}
\graphicspath{{./Fig_arXiv/}}

\begin{document}

\title{Too Far to See? Not Really! \\--- Pedestrian Detection with Scale-aware Localization Policy}

\author{Xiaowei~Zhang,
        Li~Cheng, 
        Bo~Li,
        and~Hai-Miao~Hu
\IEEEcompsocitemizethanks{\IEEEcompsocthanksitem Xiaowei~Zhang, Bo Li and Hai-Miao Hu are with School of Computer Science and Engineering, Beijing Key Laboratory of Digital Media, and State Key Laboratory of Virtual Reality Technology and Systems, Beihang University, Beijing 100191, China. E-mail: xiaowei19870119@sina.com \protect\\%
\IEEEcompsocthanksitem Li~Cheng is with Bioinformatics Institute, A*STAR, Singapore and School of Computing, National University of Singapore, Singapore.
E-mail: chengli@bii.a-star.edu.sg \protect\\%
}
}


\IEEEtitleabstractindextext{%
\begin{abstract}

A major bottleneck of pedestrian detection lies on the sharp performance deterioration in the presence of small-size pedestrians that are relatively far from the camera.
Motivated by the observation that pedestrians of disparate spatial scales exhibit distinct visual appearances, we propose in this paper an active pedestrian detector that explicitly operates over multiple-layer neuronal representations of the input still image.
More specifically, convolutional neural nets such as ResNet~\cite{HeEtAl:cvpr16} and faster R-CNNs~\cite{RenEtAl:nips15} are exploited to provide a rich and discriminative hierarchy of feature representations as well as initial pedestrian proposals. Here each pedestrian observation of distinct size could be best characterized in terms of the ResNet feature representation at a certain layer of the hierarchy; Meanwhile, initial pedestrian proposals are attained by faster R-CNNs techniques, i.e. region proposal network and follow-up region of interesting pooling layer employed right after the specific ResNet convolutional layer of interest, to produce joint predictions on the bounding-box proposals' locations and categories (i.e. pedestrian or not).
This is engaged as input to our active detector where for each initial pedestrian proposal, a sequence of coordinate transformation actions is carried out to determine its proper x-y 2D location and layer of feature representation, or eventually terminated as being background.
Empirically our approach is demonstrated to produce overall lower detection errors on widely-used benchmarks, and it works particularly well with far-scale pedestrians.
For example, compared with 60.51\% log-average miss rate of the state-of-the-art MS-CNN~\cite{CaiEtAl:eccv16} for far-scale pedestrians (those below 80 pixels in bounding-box height) of the Caltech benchmark,
the miss rate of our approach is 41.85\%, with a notable reduction of 18.68\%.
\end{abstract}

}

\maketitle
\IEEEdisplaynontitleabstractindextext

\begin{figure}[!t]
\centering
\includegraphics[width=0.79\linewidth]{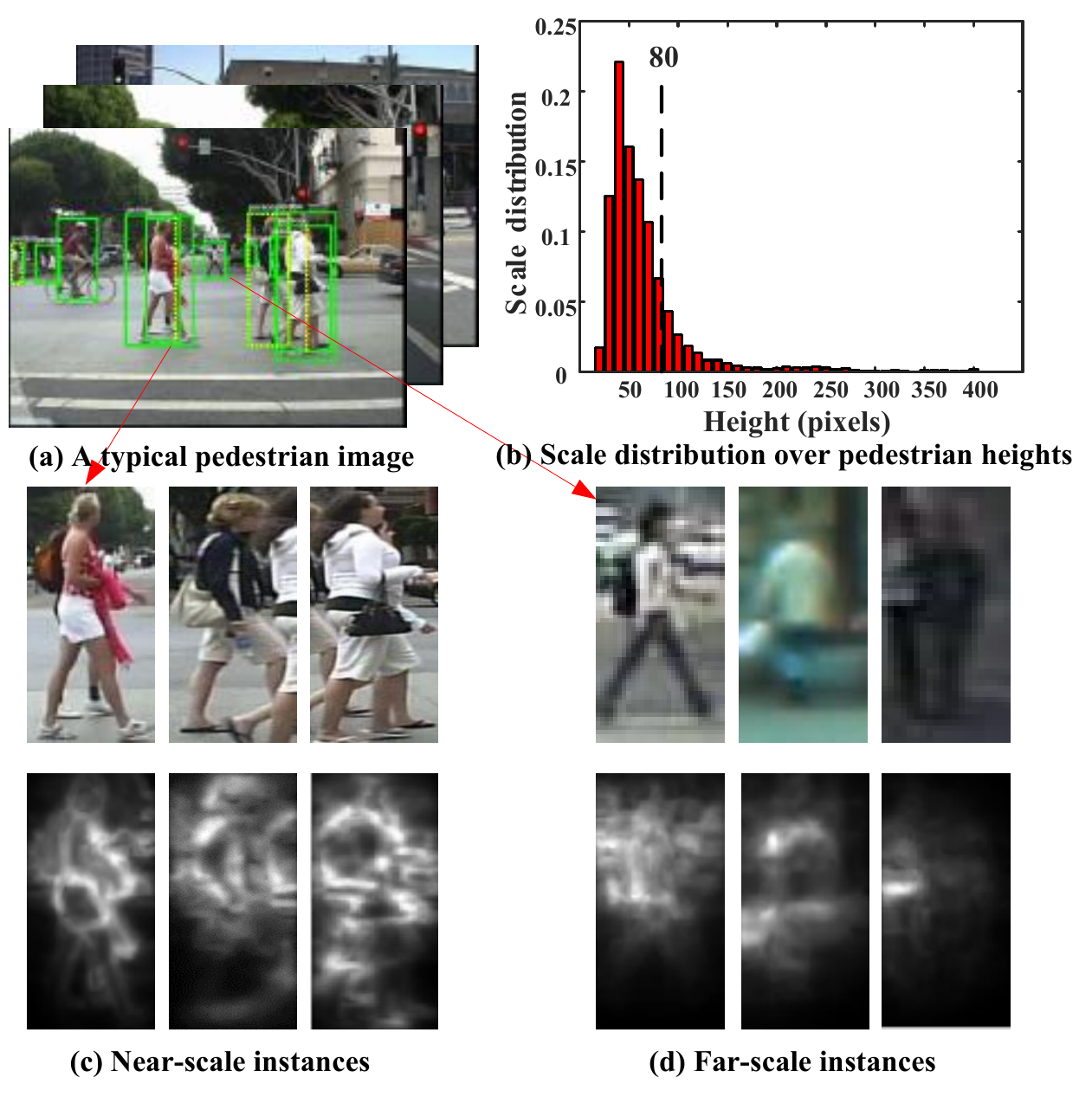}
\caption{In pedestrian detection, a typical input image usually contains multiple pedestrian instances over different scales. (a) An input image from the Caltech benchmark~\cite{DolEtAl:cvpr09}. (b) The scale distribution of pedestrian heights from the same Caltech dataset. One can observe that far-scale instances in fact dominate the distribution. (c) and (d) display exemplar visual appearance between near- and far-scale instances, as well as the corresponding neuronal feature representations from the proper layers, which are noticeably different.}
\label{fig_1}
\end{figure}

\IEEEraisesectionheading{\section{Introduction}\label{sec:introduction}}
\IEEEPARstart{P}{edestrian} detection has been an important computer vision research topic over the years, with a wide range of applications including video surveillance, intelligent vehicles, robotics, and human computer interaction, to name a few. Significant progress has been made recently~\cite{ShaEtAl:cvpr16,TiaEtAl:cvpr15,YanEtAl:iccv15,SerEtAl:cvpr13,LiEtAl:cvpr15,RedEtAl:cvpr16} in terms of both effectiveness~\cite{ParEtAl:cvpr13,PaiSheHen:iccv13,PaiSheHen:eccv14,GirEtAl:cvpr14,ZhaBenSch:cvpr15,CaoPanLi:tip16} and efficiency~\cite{DolEtAl:bmvc10,WalEtAl:cvpr10,DolAppKie:eccv12,CheEtAl:cvpr13,OuyWan:iccv13,OuyZenWan:cvpr13}. On the other hand, as presented in Fig.~\ref{fig_1}(a), a typical image often contains multiple pedestrians of different scales, and current detection performance varies significantly over scales: Although the state-of-the-art detectors typically work reasonably well with large size pedestrians where the objects are near the camera (i.e. near-scaled),
their performance becomes considerably worse when dealing with small-sized (i.e. far-scaled) ones. It has been observed that the lack of satisfactory outcome toward detecting far-scale pedestrians is due to the following inherent challenges:
First, comparing with near-scale pedestrian instances as shown in Fig.~\ref{fig_1}(c), far-scale ones often retain much less information, yet these instances contain a greater noise portion that results in obscure appearance and blurred boundaries, as displayed in Fig.~\ref{fig_1}(d). It is in general difficult to distinguish them from the background clutters. Second, for a pedestrian instance of interest, visual features are effective only at a proper scale where optimal response is obtained. The issue is more pronounced in complex scenes containing pedestrian instances of diverse scales.

Motivated by these observations, we consider in this paper a dedicated approach with a more balanced competency in detecting both near- and far-scale pedestrian instances. It possesses the following three major contributions. First, a novel active detection approach is proposed to take as input multi-layer neuronal feature representations as well as initial pedestrian proposals of the input still image, execute sequences of coordinate transformation actions to deliver the final localization (i.e. bounding-box prediction) of the pedestrian instances. These actions come from a localization policy that is learned from data by exploiting both local contextual information and multi-layer representations using deep reinforcement learning techniques. Empirical evaluation reveals that the major ingredients in our approach, namely multi-layer representations, initial pedestrian proposals, and localization policy are all important in delivering the final results. Second, our approach works with pedestrian instances across scales in a more balanced manner, and performs particularly well with far-scale instances. Empirically it is examined on several widely-used benchmarks, including Caltech~\cite{DolEtAl:cvpr09}, ETH~\cite{EssLeiGoo:iccv07}, 
and TUD-Brussels~\cite{WojWalSch:cvpr09} with consistently competitive results. For example, MS-CNN~\cite{CaiEtAl:eccv16} performs among the best on Caltech benchmark. Compared to its state-of-the-art results, the log-average miss rate of our approach is further reduced by a very significant amount of 18.68\% for far-scale pedestrians (below 80 pixels in bounding-box height), and a mild amount of 1.42\% for near-scale pedestrians (80 or more). Similar results have also been observed for the other benchmarks.
Third, our implementation and related results are made publicly available~\footnote{Our implementation, the results on benchmarks, and detailed information pertaining to the project can be found at a dedicated project webpage \url{https://web.bii.a-star.edu.sg/archive/machine_learning/Projects/objDet/pedDet/index.html}.} in support of the open-source activities of the research community.


\begin{figure*}[!t]
\centering
\includegraphics[width=0.99\linewidth]{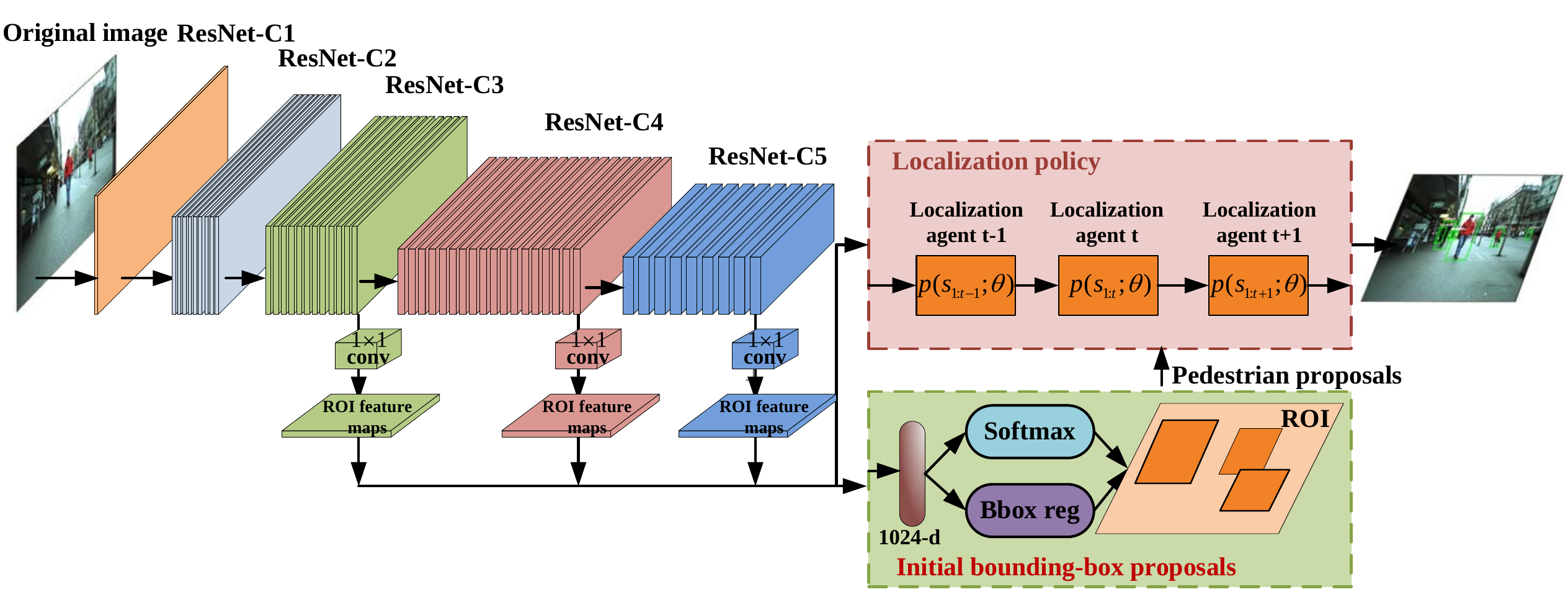}
\caption{The flowchart of our proposed approach. Multi-layer representations of ResNet are respectively utilized to compile pedestrian proposals of different sizes, which are then passed to our localization policy module to produce the final outputs.
}
\label{fig_2}
\end{figure*}

\section{Related Work}
\label{sec:related work}

There has been vast literature on pedestrian detection with lasting research activities. Due to space limit, in what follows we mainly focus on efforts that are closely related to our approach. As with many other computer vision tasks, hand-crafted features have played a critical role in attaining good performance. The histogram of oriented gradient or HOG descriptor by Dalal \& Triggs~\cite{DalTri:cvpr05} is perhaps the most well-known feature engineering technique constructed for pedestrian detection. This is extended to the integral channel features (ICF) by Dollar et al.~\cite{DolEtAl:bmvc09} to efficiently extract features such as local sums, histograms, and Haar features using integral images, which is further improved in several ways by ACF~\cite{DolEtAl:tpami14}.

Recently, Convolutional Neural Network (CNN) based methods~\cite{TiaEtAl:cvpr15,TiaEtAl:iccv15,HosEtAl:cvpr15,CaiSabVas:iccv15,ZhaEtAl:eccv16} have made significant progresses in pedestrian detection, among others. An unsupervised method based on convolutional sparse coding is used in Sermanet et al.~\cite{SerEtAl:cvpr13} to pre-train CNN for pedestrian detection. Tian et al.~\cite{TiaEtAl:cvpr15} jointly optimize pedestrian detection with other semantic tasks including pedestrian and scene attributes. Yang et al.~\cite{YanChoLin:cvpr16} investigate the augmentation of CNN with scale-dependent pooling and layer-wise cascaded rejection classifiers to detect objects efficiently. Cai et al.~\cite{CaiSabVas:iccv15} introduce complexity-aware cascaded detectors by leveraging both hand-crafted and CNN features for an optimal trade-off between accuracy and speed. The RCNN developed by Girshick et al.~\cite{GirEtAl:cvpr14} nicely integrates an object proposal mechanism within CNN framework. This leads to SPPnet~\cite{HeEtAl:eccv14} that improves the detection speed of RCNN by evaluating CNN features once per image. Built on top of RCNN and SPPnet, Fast-RCNN~\cite{Gir:iccv15} brings up the idea of single-stage training and multi-task learning of both a classifier and a bounding box regressor. Furthermore, Faster-RCNN~\cite{RenEtAl:nips15} considers a region proposal network that shares full-image convolutional features with the detection network to efficiently predict object proposals, instead of the time-consuming techniques such as selective search, leading to a significant speedup for proposal generation.

Multi-layer approaches have also been developed for detecting objects across multiple scales. \cite{CaiEtAl:eccv16,HeEtAl:eccv14} attempt to alleviate the inconsistency of RPN~\cite{RenEtAl:nips15} by employing an upsampling operation on inputs at both training and testing stages, which nevertheless is significantly more demanding in terms of computation and memory. SA-FastRCNN~\cite{LiEtAl:cvpr15} develops a divide-and-conquer strategy based on Fast-RCNN that uses multiple built-in subnetworks to adaptively detect pedestrians across scales. Similarly, MS-CNN~\cite{CaiEtAl:eccv16} works with multiple layers to match objects of different scales. SSD~\cite{LiuEtAl:eccv16} discretizes the output space of bounding boxes into a set of template boxes over varying aspect ratios and scales. Complementary detectors are then utilized at different output layers that collectively give rise to a strong multi-scale detector.
A strategy adopted by some of these methods~\cite{GirEtAl:cvpr14,DolEtAl:bmvc10} is to train a single classifier at a fixed resolution, meanwhile the input image is rescaled to several distinct sizes, and the associated features are re-computed as independent inputs. By aiming to attain such scale-invariancy property, empirically this strategy produces improved detection results, at a price of being computationally very costly.
Another strategy adopted by~\cite{BenEtAl:cvpr12,FelEtAl:tpami10,SerEtAl:cvpr13,EnzGav:tip11} is to simultaneously engage multiple object detectors to a single resolution of the input image, with each focusing on a unique local patch size. This strategy avoids the repeated computation of feature maps and tends to be more efficient, nonetheless, it is often difficult to sufficiently represent these feature responses within a single resolution.
More recent works such as Faster-RCNN~\cite{RenEtAl:nips15} address the issue with a multi-layer region proposal network (RPN), which achieves excellent object detection performance. However, RPN generates multi-layer proposals by sliding a fixed set of filters over a fixed set of convolutional feature maps. There thus exist potential mismatches between the sizes of objects and filter receptive fields, as the object resolutions are variable, yet the sizes of filter receptive fields are fixed.

In practice, it sometimes leads to compromised results with particularly poor detections for far-scale objects.

Instead we consider a different strategy as follows.
By engaging multiple convolutional feature layers, following the representation theory, outputs of higher layers encode semantic information of targets that are robust to significant appearance variations. However, their spatial resolutions tend to be too coarse to precisely localize the far-scale pedestrian instances. On the flip side, outputs of the lower convolutional layers provide more precise localization, but are less invariant to the objects appearance changes. This motivates us to consider learning a localization policy to actively identify the suitable representation layer as well as the appropriate filter size for the object of interest.
Meanwhile, deep reinforcement learning techniques have been considered recently by~\cite{CaiLaz:iccv15,JieEtAl:NIPS16} to learn a localization policy that identifies the bounding-box of an object of interest, as well as by~\cite{MnihEtAl:nips14} for digit recognition with a recurrent attention mechanism.
Yet the impact of multiple pedestrians of different scales as in our scenarios are not considered.

In the meantime, thorough surveys on pedestrian detection have been conducted by e.g.~\cite{EnzGav:tpami09,DolEtAl:tpami12}.
The evaluations in~\cite{DolEtAl:tpami12} is more recent, where the pedestrian instances are geometrically grouped into three scales: near, medium, and far. \cite{DolEtAl:tpami12} proceeds to present their Caltech dataset as a reasonable sample of typical pedestrian images, where the data is collected by adopting a standard vehicle-mounted camera. In this influential dataset, near-scale corresponds to a pedestrian object of 80 or more pixels in height, medium-scale is for a pedestrian object between 30-80 pixels, and far-scale is for 30 pixels or less. In terms of physical distance between the pedestrian instance and the camera, these three scales (i.e. near, medium, far) roughly correspond to a typical pedestrian with a distance below 20, 20-60, and over 60 meters, respectively. As noted in~\cite{DolEtAl:tpami12}, the medium- and far-scale instances comprises the majority (84\% in this case) of the pedestrian population in a typical imaging setting, and human observers have no trouble in detecting most of them. In contrast, existing pedestrian detection methods perform rather poorly in the presence of these instances, despite the fact that the same detectors work very well with those near-scale instances.
Take one latest effort,MS-CNN~\cite{CaiEtAl:eccv16} for example, it has been reported that empirically their detector is capable of achieving 3.30\% log-average miss rate for near-scale pedestrians in Caltech Pedestrian Benchmark~\cite{DolEtAl:cvpr09}, the error rate however increases to 60.51\% for medium- and far-scale pedestrians.
Similar observations have also been expressed in other works~\cite{HoiChoDai:eccv12,ZhaEtAl:cvpr16}.
It inspires us to consider a simpler grouping of pedestrian instances into two scales: near-scale vs. far-scale, which corresponds to near vs. medium \& far in the Caltech benchmark. The focus of this paper is to achieve a more balanced detection performance for both near- and far-scale pedestrian instances.

\section{Our Approach}
\label{sec:section 3}

We start by laying down some necessary notations.
A bounding-box (bbox) coordinate can be identified by $\vec{b}=(b^x,b^y,b^w,b^h) \in \mathbb{R}^4$ for its top-left corner $(b^x,b^y)$, width, and height.
At the same time, let $p=p_{\vec{b}} \in \{0, 1\}$ be the probability of whether the bbox is a pedestrian bbox.
An input pedestrian instance is denoted as $\vec{x}$, also referred to as an anchor, and its label is $\vec{y}=\left(p, \vec{b}\right)$.
Here $\vec{x}$ corresponds to all the information (such as contextual information, specific feature layer representation) in the input image that is related to the pedestrian instance,
which could be obtained by for example applying a sliding window. At any time,
if the intersection-over-union (IoU) of the current sliding window (anchor) and a closest ground-truth pedestrian bbox is above a relatively high threshold (say 0.5),
then its label $\vec{y}$ is assigned as $p=1$ and $\vec{b}$ being the ground-truth bbox.
Otherwise, if the IoU is below a relatively low threshold (say 0.3) with any ground-truth pedestrian bboxes,
then its label becomes $p=0$ and $\vec{b}$ being zero-valued.
We further denote a predicted output of $\vec{x}$ as $\hat{\vec{y}}=\left(\hat{p}, \hat{\vec{b}}\right)$, with $\hat{p}$ being the objectness score, i.e. the probability of being a pedestrian bbox.
The training set is composed by a set of (non-)pedestrian examples $\mathcal{D} = \left\{ \vec{x}_i, \vec{y}_i \right\}_{i=1}^D$,
which could be obtained by running the sliding window over multiple representation layers of the annotated pedestrian training images.
To explicitly account for the fact that these pedestrian instances are at different scales, denote as $M$ the number of neural representation layers (here $M$=3),
and the training set is partitioned to $\mathcal{D}= \mathcal{D}^1 \cup \mathcal{D}^2 \cup \ldots \cup \mathcal{D}^M$ and $\mathcal{D}^m \cap \mathcal{D}^n = \varnothing$  $\forall m, n$,
with each disjoint subset $\mathcal{D}^m$ containing specifically the training examples at current layer.

A high-level overview of our approach is displayed in Fig.~\ref{fig_2}, which consists of two main stages: An input image will first pass through an initialization stage to make ready the multi-layer feature representations and the initial pedestrian proposals, they are then fed into an active detection stage where a dedicated localization policy is engaged to produce the final bbox predictions $\left\{ \hat{\vec{y}}_i \right\}$ by executing sequences of coordinate transformation actions.

\begin{figure}[!t]
\centering
\includegraphics[width=0.89\linewidth]{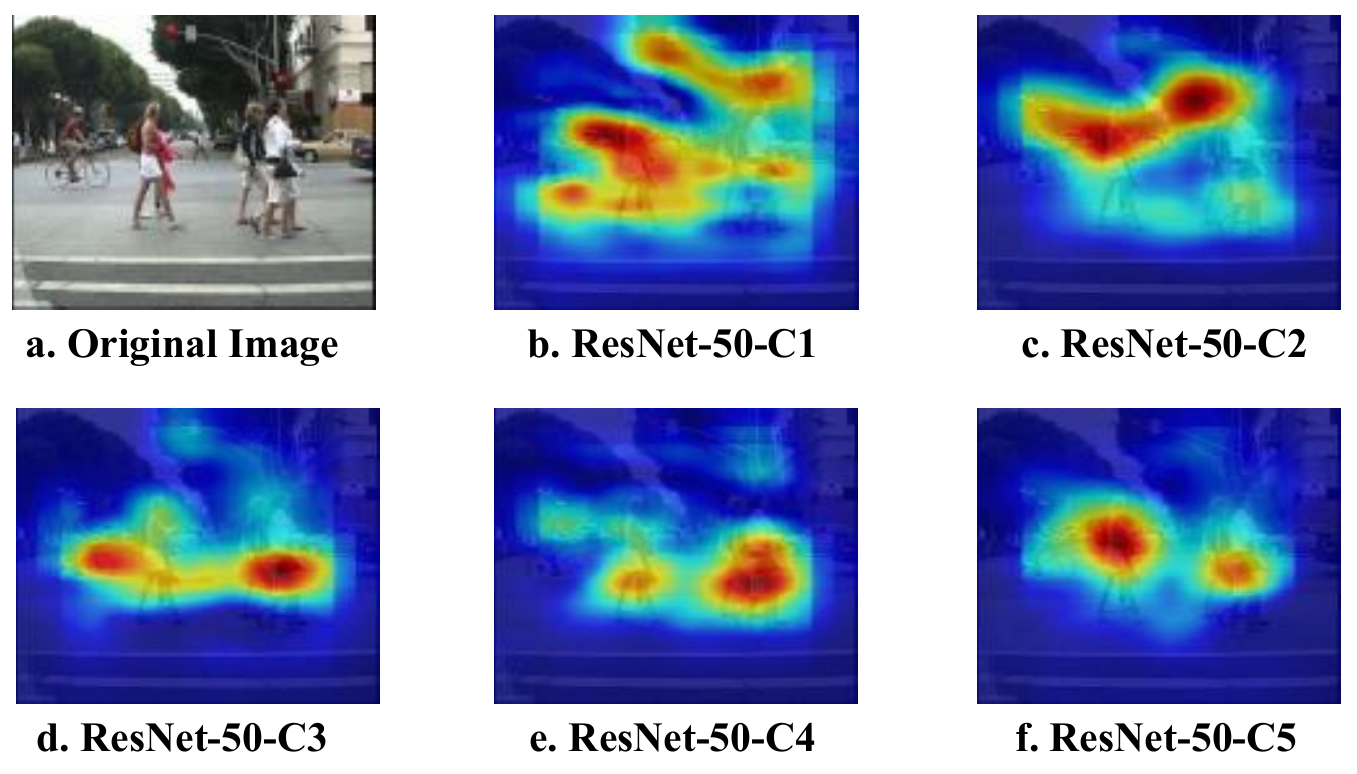}
\caption{Visualization of the pedestrian proposal activations across multiple neural representation layers. See text for more details.}
\label{fig_3}
\end{figure}

\begin{figure}[!t]
\centering
\includegraphics[width=0.98\linewidth]{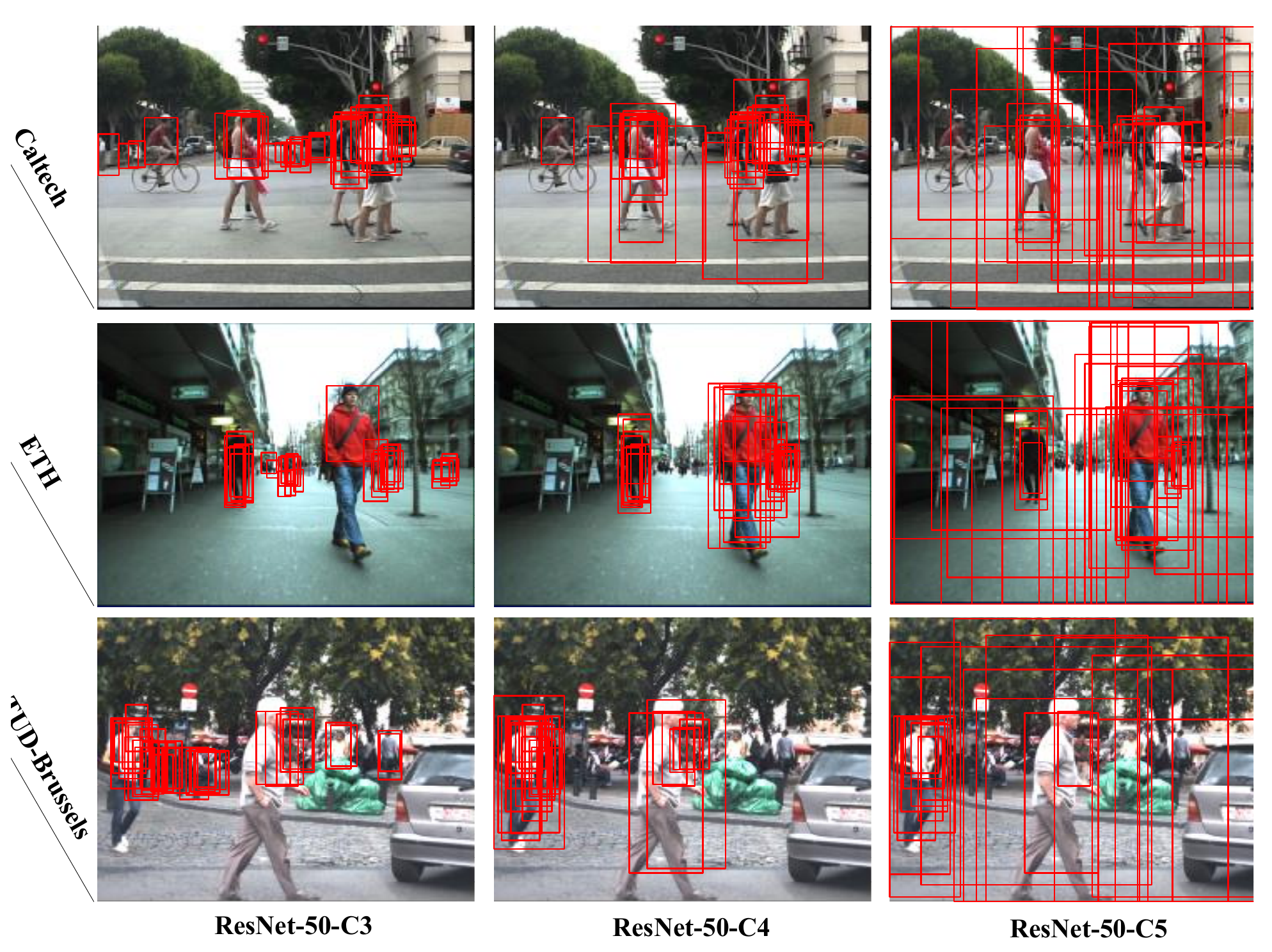}
\caption{A visual elucidation on the effect range of our resultant initial proposals at each layer of the feature representation hierarchy.}
\label{fig_4}
\end{figure}

\subsection{Multi-layer Representation and Initial Pedestrian Proposals}
\label{sec:section 3}

It has been shown in Fig.~\ref{fig_1} that given multiple pedestrian instances of different scales being presented in a typical image, their best responses usually occur at distinct feature representation layers. This is revealed in more details at Fig.~\ref{fig_2}, where three intermediate convolutional layers of ResNet~\cite{HeEtAl:cvpr16}, i.e. ResNet-50-C3, ResNet-50-C4 and ResNet-50-C5, are employed as our multi-layer feature representations.
Moreover, Fig.~\ref{fig_2} provides more detailed information over the first five ResNet layers (i.e. ResNet-50-C1 to ResNet-50-C5).
Here the class activation maps of~\cite{ZhouAl:cvpr16} are applied to visualize the responses at different convolutional layers for generating pedestrian proposals.
In general, lower representation layers may have a strong activation of convolutional neurons for far-scale pedestrians. Similar phenomenon for near-scale pedestrian instances usually emerge at higher layers.
Moreover, higher layer neurons tend to encode more global and semantic information of objects that could be robust against significant appearance variations, while their spatial resolutions tend to be overly coarse, thus unable to accurately locate far-scale pedestrian instances. On the other hand, outputs of lower convolutional layers provide more precise localization, while being less invariant to objects' appearance changes.
It can be seen from Fig.~\ref{fig_3} that the first two layers namely ResNet-50-C1 and ResNet-50-C2 could be saturated by the amount of false alarms from backgrounds, while ResNet-50-C3 seems a satisfactory starting layer for effective multi-layer representation.
For near-scale objects such as the one at the center location of Fig.~\ref{fig_3}, their feature responses appear across multiple layers, yet the last two layers and especially the very last layer here seems to retain best responses. The feature responses from the rest (i.e. the first two) layers are still noticeable but with more unnecessary local or even background details due to small sized receptive fields from lower-layers. This is further illustrated in Fig.~\ref{fig_4}, where far-scale pedestrians are most comfortably picked up at lower layers, and near-scale ones are largely detected at upper layers. Meanwhile, it appears to be difficult to identify far-scale ones at upper layers, and vice versa.
In practice, three layers namely ResNet-50-C3, ResNet-50-C4, and ResNet-50-C5, are chosen as the working scale space in this paper.

As displayed in Fig.~\ref{fig_2}, an input image is passed through the ResNet layers to form multi-layer feature representations,
where bbox proposals of different scales are also generated.
The region of interest (RoI) pooling layer of~\cite{YanChoLin:cvpr16} is then utilized to pool the feature maps of each pedestrian proposal into a fixed-length feature vector,
which is fed into a fully connected layer.
At last, each pedestrian proposal $\hat{\vec{y}}$ ends up with two outputs, one accounts for classification score $\hat{p}$, and the other one predicts the bbox position $\hat{\vec{b}}$.

Up to now it should be clear that this module serves the purpose of providing a multi-layer feature representation of the input image.
We then aim to produce predictions on both initial bbox locations (pedestrian proposals) and the associated objectness score, following that of Faster R-CNNs~\cite{RenEtAl:nips15}.
Training this module involves estimating the weight parameters of the pedestrian proposal network as depicted in the green box of Fig.~\ref{fig_2}.
The objective function $L_{\mathcal{D}} \left(\mathcal{W} \right)$ is thus a weight parameterized neural network function defined over our training examples $\mathcal{D}$,
where $\mathcal{W}$ stands for the weight parameters.
As it contains multi-layer feature representations as input, let $l^m \left( y_i, \hat{y}_i | \mathcal{W} \right)$ denote the loss function of the $i$-th training example at the $m$-th layer of the Resnet feature representation, and without loss of generality, denote by $\mathcal{M}=\{3,4,5\}$ the aforementioned Resnet layers considered in our paper. The objective function is then formulated as

\begin{align}
\label{eq1}
L_{\mathcal{D}}(\mathcal{W})=\sum\limits_{m \in \mathcal{M}} \sum\limits_{i\in{\mathcal{D}^m}} \alpha_m l^m( y_i, \hat{y}_i|\mathcal{W}),
\end{align}
where
$\alpha_m$ is the weight of current loss $l^m( y_i, \hat{y}_i|\mathcal{W})$ that can be computed by the softmax function $\alpha_m = \frac{e^{\hat{\alpha}_m}} {\sum\limits_{m'=1}^M e^{\hat{\alpha}_{m'}}}$, with $\hat{\alpha}_m = \frac{1} {(1+e^{- \frac{h-\bar{h}_m}{\gamma_m} })}$.
$\bar{h}_m$ denoting the average height of the pedestrians from the $m$-th layer in the training set.
For example, the respective $\bar{h}_m$ values in Caltech benchmark are 48, 96, and 156.
${\gamma}_m$ is a scaling factor, which is empirically set to 5, 20, and 10, respectively for the three layers.
Note a smaller ${\gamma}_m$ value leads to larger gap between the weights of pedestrian instances from different layers.

Furthermore, our multi-task loss function $l^m( y_i, \hat{y}_i|\mathcal{W})$ on the $i$-th example at layer $m$ is defined as:
\begin{align}
\label{eq_lm}
l^{m} \left( y_i, \hat{y}_i|\mathcal{W} \right) = l_{\mathrm{cls}} \left( p_i,\hat{p}_i \right) + \lambda p_i l_{\mathrm{loc}} \left(\vec{b}_i,\hat{b}_i \right),
\end{align}
where $p_i$ is 1 if the anchor is labeled positive, and 0 otherwise. $\hat{p}_i$ denotes the predicted probability of the anchor being a pedestrian proposal.
$\vec{b}_i=(b_i^x,b_i^y,b_i^w,b_i^h)$ represents the ground-truth bbox associated with a positive anchor,
and $\hat{\vec{b}}_i=(\hat{b}_i^x,\hat{b}_i^y,\hat{b}_i^w,\hat{b}_i^h)$ denotes the predicted bbox.
The second loss term is defined on the regression task, for which we adopt the same smooth-L1 loss function of Faster R-CNN~\cite{RenEtAl:nips15},
that is, $l_{\mathrm{loc}}=R \left(\vec{b}_i-\hat{\vec{b}}_i \right)$. Here $R(\vec{b})=0.5 \|\vec{b}\|^2$ if $\|\vec{b}\|<1$, and $R(\vec{b})=\|\vec{b}\|-0.5$ otherwise.
The term $p_i l_{\mathrm{loc}}$  is to activate the regression loss only for positive anchors ($p_i=1$), which is disabled otherwise ($p_i=0$).
$\lambda$ is a trade-off parameter (empirically set to 10) for which a larger value places a stronger emphasis on good bbox locations.
In terms of the first loss, i.e. the classification loss, it takes the form of log-loss $l_{\mathrm{cls}}=-\sum_i \log \hat{p}_i$. 
In practice a variation of this form is adopted, as to be described next.

Consider the training subset of a specific representation layer, which can be further decomposed as $\mathcal{D}^m=\{\mathcal{D}_+^m, \mathcal{D}_-^m \}$.
Our bbox proposals can be generated at sliding-window locations that are partitioned into positive examples $\mathcal{D}_+^m$ and negative ones $\mathcal{D}_-^m$.
An anchor is centered at the sliding window on each layer associated with aspect ratio of 0.41 (width to height, adopted based on~\cite{DolEtAl:cvpr09}).
For positive examples $\mathcal{D}_+^m$ , two kinds of anchors are labeled positive:
(i) The anchor that has an IoU overlap higher than 0.5 with any ground-truth bbox; (ii) The anchor that has the highest IoU overlap with a ground-truth bbox.
Note in the first case, a single ground-truth bbox may assign positive labels to multiple anchors;
The second case is to ensure that at least one positive example exists for any ground-truth bbox.  For negative examples $\mathcal{D}_-^m$ , a negative label is assigned if its IoU ratio is lower than 0.3 over all ground-truth boxes. The rest anchors that are neither positive nor negative are simply ignored as having no contribution to the training objective.
In addition, to compensate for the imbalance of the positive and negative examples,
a small number of negative examples are first uniformly sampled.
The training set is then bootstrapped by ranking and sampling the negative examples according to their objectness scores to obtain hard negative examples.
As potentially there are far more negative candidates, a balancing constant $\gamma\geq 1$ is used
such that the negative set is constrained by $\left| \mathcal{D}_-^m \right|=\gamma \left| \mathcal{D}_+^m \right|$.

Finally, by considering contributions of both positive and negative examples, the classification loss term now becomes a weighted cross-entropy:
\begin{small}
\begin{align}
\label{eq_l_cls}
l_{\mathrm{cls}}=\frac{1}{1+\gamma}\frac{1}{\left| \mathcal{D}_+^m \right|} \sum\limits_{i\in{\mathcal{D}_+^m}}-\log \hat{p_i} + \frac{\gamma}{1+\gamma}\frac{1}{\left| \mathcal{D}_-^m \right|} \sum\limits_{j\in{\mathcal{D}_-^m}} \log \hat{p_j}.
\end{align}
\end{small}

\begin{figure}[!t]
\centering
\includegraphics[width=3.0in]{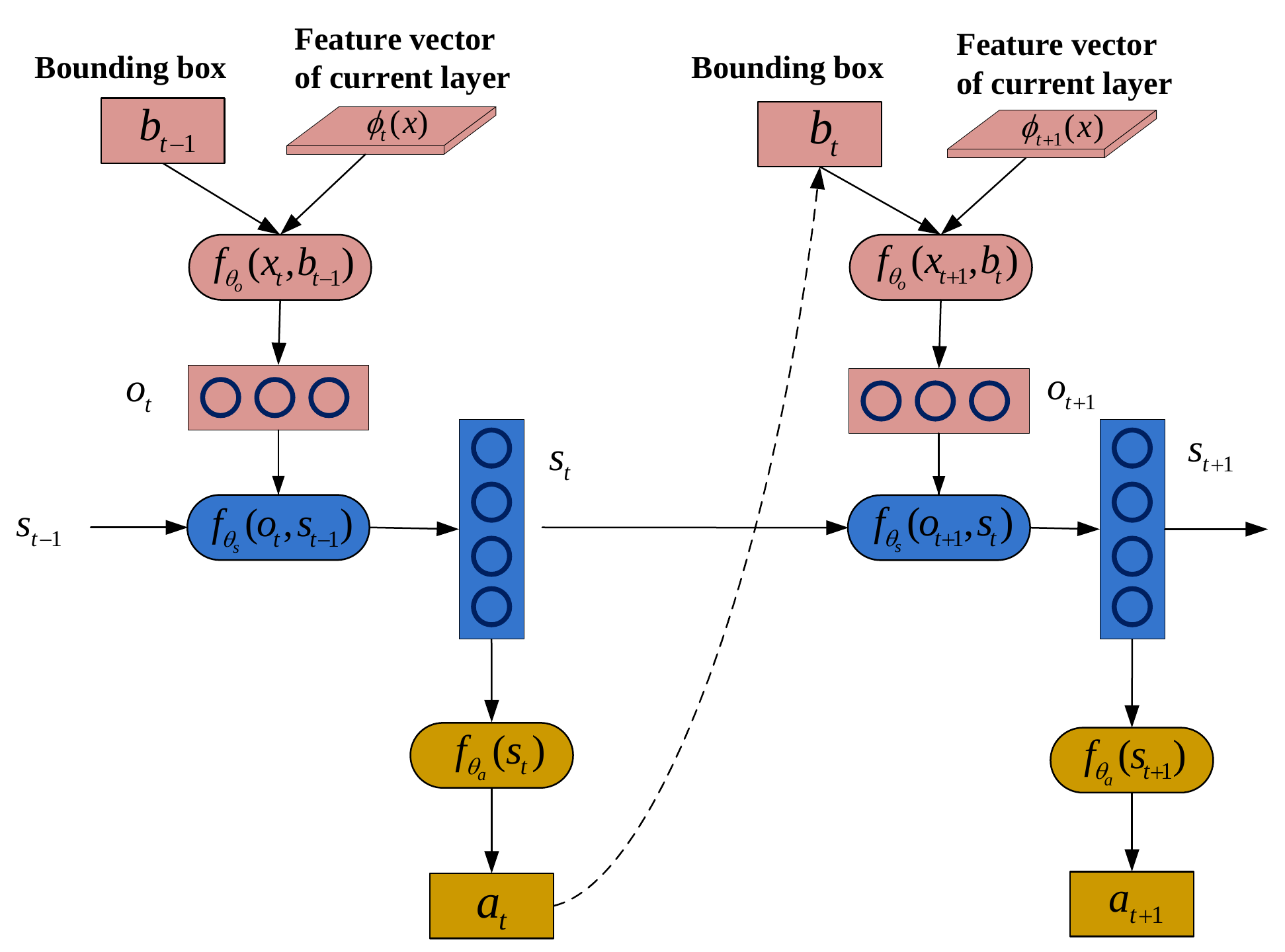}
\caption{A schematic illustration of our active detection module in the process of producing a series of transformation action $\left( \ldots, a_t, a_{t+1}, \ldots \right)$, when working with a possible pedestrian instance $\vec{x}$ of an input image. See text for details.}
\label{fig_5}
\end{figure}

\begin{figure}[!t]
\centering
\includegraphics[width=0.69\linewidth]{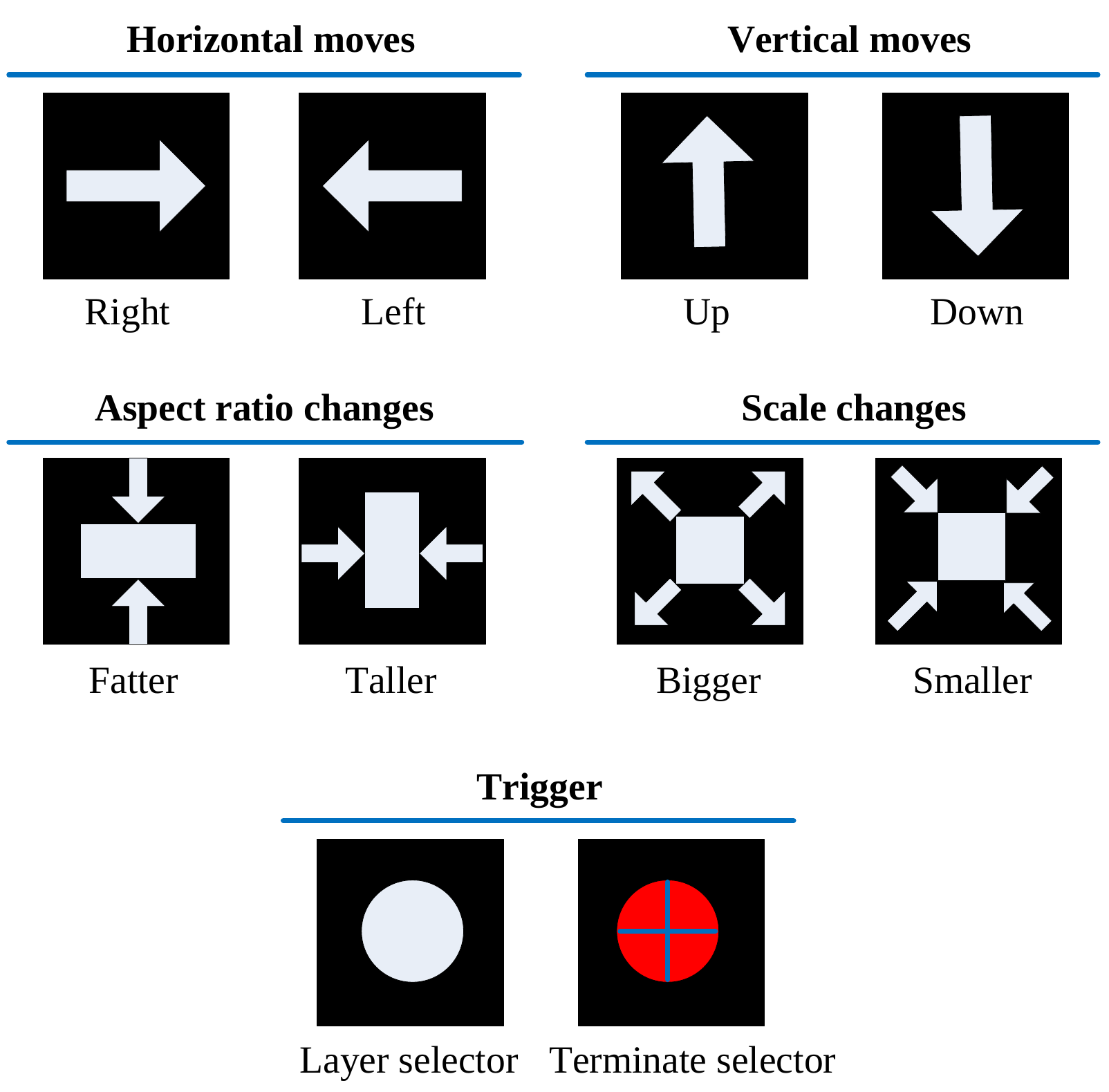}
\caption{The set of actions considered in our localization policy.}
\label{fig_6}
\end{figure}

\subsection{Our Active Detector Model}

In conjunction with the initial proposals described in the previous section,
an active detection model is introduced for improving pedestrian bbox predictions by performing a series of coordinate transformation actions.
It allows an agent to adaptively choose the feature maps of proper resolutions catering for current pedestrian distances from our multi-layer feature representation.
Inspired by the recent work of~\cite{MnihEtAl:nips14} for digit recognition, we consider a recurrent neural network (RNN) based model,
where at each time step $t$, both spatial and temporal contextual information are incorporated to decide current transformation action.

Fig.~\ref{fig_5} displays the sequential process of applying our active detector model to a possible pedestrian instance $\vec{x}$ of an input image.
At start time $t:=1$, let $\vec{b}_{t-1}$  be the initial bbox proposal and $\phi_t \left(\vec{x}\right)$ be the feature representation of its corresponding layer.
A finite sequence of bbox transformation actions $\left(a_1, a_2, \ldots \right)$ is then executed based on a recurrent neural network which is trained by reinforcement learning. Note the size of the feature vector $\phi_t \left(\vec{x}\right)$ may vary due to multi-layer representation.
Specifically, the C3 to C5 layers of ResNet are employed in this paper, which has different number of feature maps of size 256, 512, and 1024, respectively. Moreover, a spatial pixel location in C3 to C5 layers roughly corresponds to a patch size of 32$\times$32, 64$\times$64, and 128$\times$128 in the original image, respectively. So a tiny pedestrian might simply be overlooked. To address this issue, we introduce a lower bound size of 4$\times$4 over all these three representation layers, which is enforced if the size of the object of interest is less than 4$\times$4. This way the contextual information is incorporated especially for far-scale pedestrian instances. The approach is followed by a resize operation where larger objects (i.e. with sizes exceeding 4$\times$4) are rescaled with bicubic interpolation to 4$\times$4. Thus the feature vector dimensions of these representation layers become $4096=4\times4\times256$, 8192, and 16384, respectively.
Now, at any time $t \in \left\{1,\ldots, T\right\}$, the following neural net functions are engaged:
First, a succinct description of this instance distilled from $m$-th layer of its ResNet feature representation is provided by a fully connected (fc) layer and a follow-up vector valued ReLU activation function, as
\begin{align}
\label{eq_ot}
\vec{o}_t := f_{\theta_{\vec{o}}}\left( \vec{b}_{t-1}, \vec{x} \right) = \max \left( \theta_{\vec{o}}^{(m)} \phi_t \left(\vec{x}\right), 0 \right),
\end{align}
where $m = m(t) \in \mathcal{M}$ denotes the particular layer of interest as considered in $\phi_t \left(\vec{x}\right)$ at time $t$.
Second, an internal information state of $\vec{s}_t$ at time $t$ is maintained that summarizes the information extracted from the history of past observations $\vec{s}_{t-1}$ and the current environment:
\begin{align}
\label{eq_st}
\vec{s}_t := f_{\theta_{\vec{s}}}\left( \vec{o}_t, \vec{s}_{t-1} \right) = \tanh \left( \theta_{\vec{s},1} \vec{o}_t + \theta_{\vec{s},1} \vec{s}_{t-1} \right),
\end{align}
with $\theta_{\vec{s}} = \left\{ \theta_{\vec{s},1}, \theta_{\vec{s},2} \right\}$. It coincides with the cell state signals of the LSTM units.
Finally, a transformation action $a_t$ is obtained by randomly drawing from the softmax distribution conditioned on the current state
\begin{align}
\label{eq6}
f_{\theta_a}\left( \vec{s}_t \right) := \mathrm{softmax} \left( \theta_{a} \vec{s}_{t} \right).
\end{align}
Empirically in our paper, $\vec{o}_t$ is set to be a 1024-dim vector, $\vec{s}_t \in \mathbb{R}^{64}$ by considering 64-unit LSTM for our RNN. Subsequently the set of neural net function parameters $\mathcal{\theta}:= \left\{ \theta_{\vec{o}}, \theta_{\vec{s}}, \theta_{a} \right\}$ are of the following dimensions: For $\theta_{\vec{o}}$ of Eq.~\eqref{eq_ot}, its size is 4096 (or 8192 or 16384) $\times$ 1024, for layer $m=$3 (or 4 or 5). For $\theta_{\vec{s}}$ of Eq.\eqref{eq_st}, the dimension is 64$\times$1024 for $\theta_{\vec{s},1}$ and 64$\times$64 for $\theta_{\vec{s},2}$. For $\theta_a$ of Eq.\eqref{eq6}, its size is 10$\times$64, since there are 10 distinct action types considered in our work as to be described soon.

\begin{figure}[!t]
\centering
\includegraphics[scale=0.5]{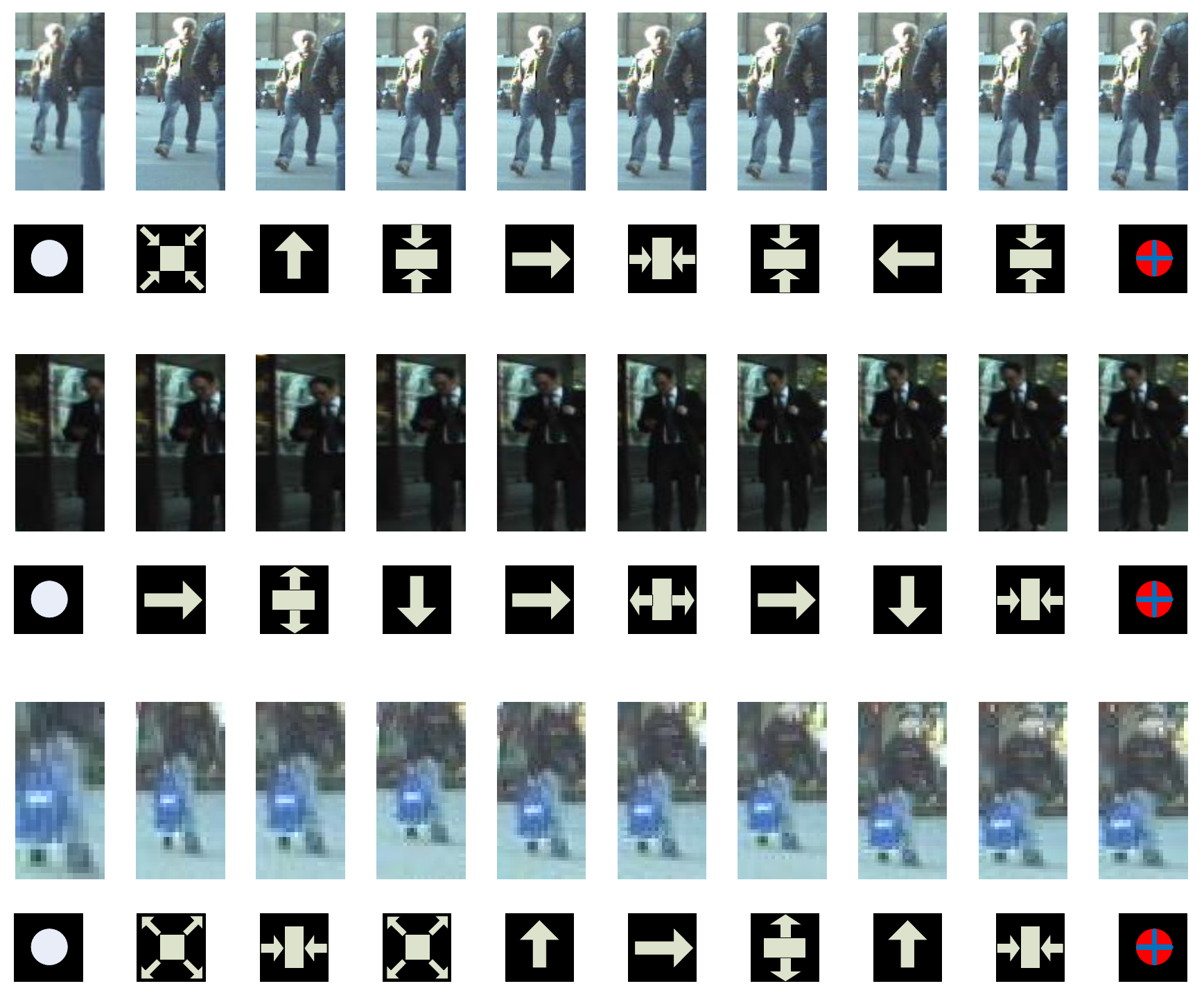}\\
\caption{Exemplar action sequences executed by our localization policy.}
\label{fig_67}
\end{figure}

\begin{figure}[!t]
\centering
\includegraphics[width=0.69\linewidth]{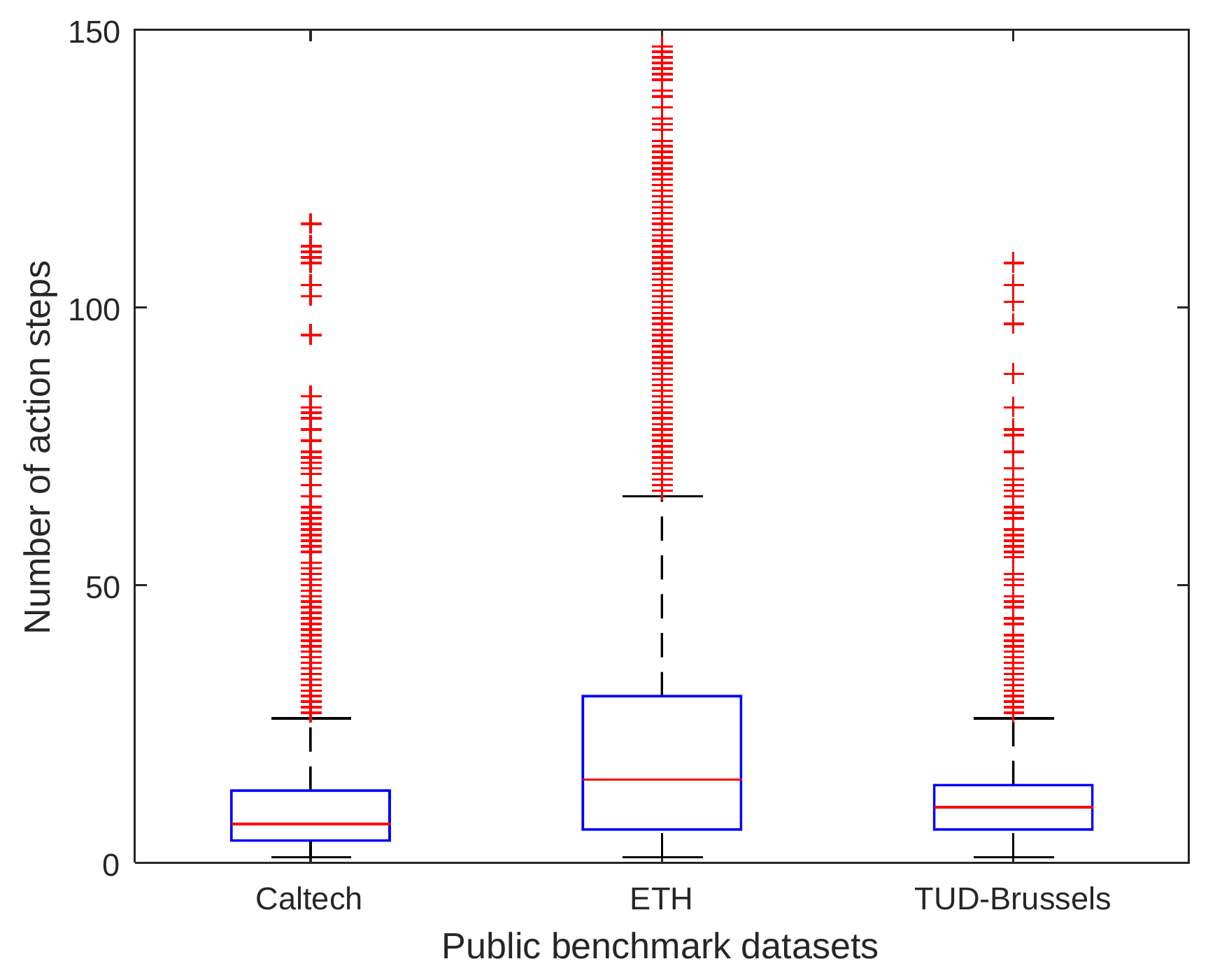}\\
\caption{Box-plot illustration of the distributions of action sequence length over different benchmark datasets.}
\label{fig_numOfActions}
\end{figure}

At this moment, consider a Markov decision process or MDP~\cite{SutBar:book98} that comes with a set of information states $\mathcal{S}$, a set of coordinate transformation actions $\mathcal{A}$,
and a reward function $r$.
At a time step $t$, the agent observes the state $s_t \in \mathcal{S}$ that contains sufficient information characterizing the current bbox prediction, its history,
its image patch and contextual features, and proceeds to decide on a specific action $a_t \in \mathcal{A}$.
When terminated, the finite sequence of actions is collectively evaluated by reward $r:=r_t$.
several such visual examples of the action sequences are shown in Fig~\ref{fig_67}.
Note that the sequence lengths and the first action in these examples of Fig~\ref{fig_67} happen to be the same, while in general they could be very different.
The box-plots in Fig.~\ref{fig_numOfActions} summarize the distribution of action sequence length over three different benchmark datasets.
The medium action sequence length is around 10, that is, 10 consecutive actions from an initial bbox prediction to reach the final decision.
Training the localization agent involves maximizing the expected discounted sum of rewards, and in practice we consider the REINFORCE~\cite{Wil:mlj01} method.
In what follows, we will delineate the concrete details of the involved MDP.

\textbf{State:}
At time $t$, our localization agent resides at an information state $s_t \in \mathcal{S}$ that encodes the current bbox prediction and the corresponding image patch features,
as well as the contextual knowledge including a description of the bbox prediction history and the related circumstantial multi-layer feature representations. This is compactly represented as the 64-dim vector $\vec{s}_t$.

\textbf{Action}: Fig.~\ref{fig_6} enumerates the set of ten actions $\mathcal{A}$ organized into two categories:  transformation actions and trigger actions, as considered in this work.
At time $t$, our agent decides on a specific action $a_t \in \mathcal{A}$ drawn from a distribution function of Eq.~\eqref{eq6}, which lands on a subsequent bbox prediction $\vec{b}_{t+1}$:
If it is a horizontal move, we have $b^x_{t+1} = b^x_{t} + d$ with $d \in \mathbb{R}$ being negative or positive, which corresponds to move leftward or rightward respectively;
Similarly, for a vertical move, $b^y_{t+1} = b^y_{t} + d$ with $d<0$ or $d>0$ for moving upward or downward, respectively;
If it is an aspect ratio change action, we have either $b^h_{t+1} = b^h_{t} + d$ for changing the height or $b^w_{t+1} = b^w_{t} + d$ for changing the width;
A scale change action amounts to modify the bbox size,
which is described by both $b^h_{t+1} = b^h_{t} \times d$ and $b^w_{t+1} = b^w_{t} \times d$, with $d \in (0,1)$ for shrinking or $d>1$ for scaling up.
Finally, the layer trigger allows an opportunity to explore the representation hierarchy for optimal pedestrian scale,
while the terminate trigger is to stop the sequence of coordinate transformation actions.

\textbf{Reward}: The reward function $r:=r_T$ at final time point $T$ serves the purpose of encouraging to learn localization policies that can detect pedestrians with high performance. Consequently a more strict detection criteria is enforced internally:
The reward is set to $r = 1$ if the IoU between the finally predicted pedestrian bbox and its best-matched ground-truth bbox is above 0.7; It is set to $r=0$ if the IoU is below 0.2, for any IoU value in-between, the reward is set to be the IoU value.

\begin{figure}[!t]
\centering
\subfloat[Near-scale (height$\geq$80 pixels)]{
\includegraphics[width=0.49\linewidth]{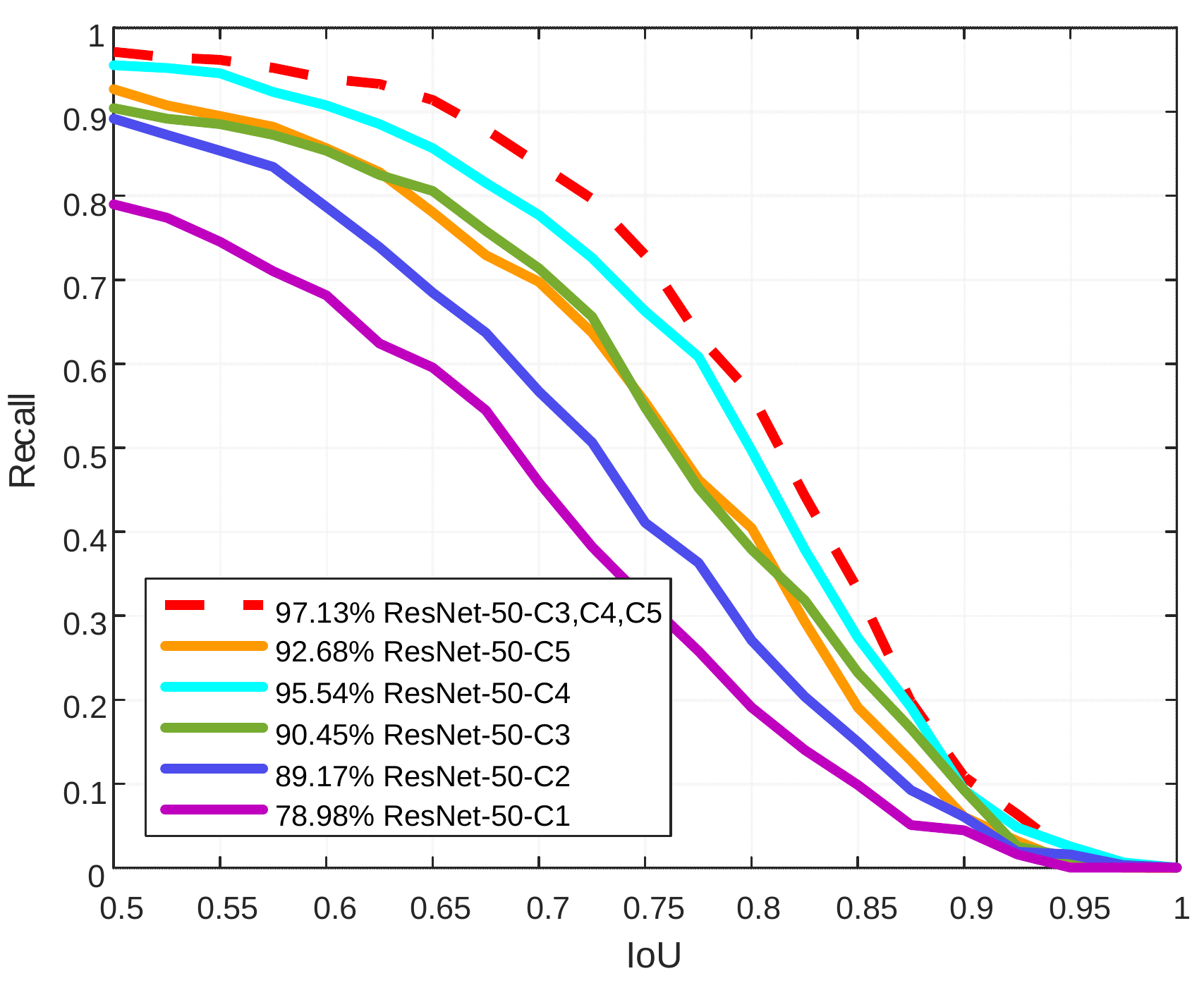}
}
\subfloat[Far-scale (height$<$80 pixels)]{
\includegraphics[width=0.49\linewidth]{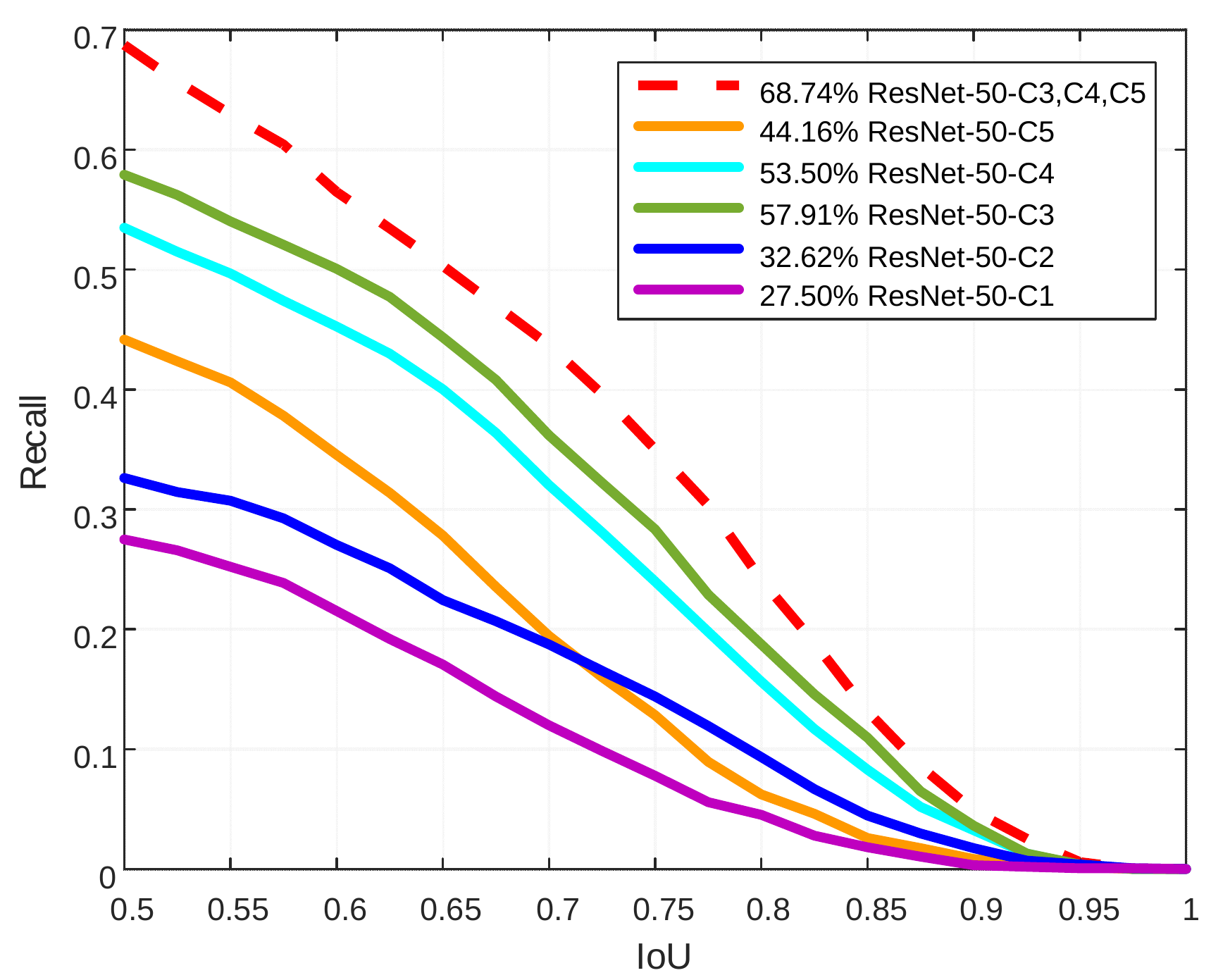}
}
\caption{Recall rate of near- vs. far-scale pedestrians with varying intersection-over-union scores (IoUs) at different Resnet representation layers (ResNet-50-C1 to ResNet-50-C5).}
\label{fig_7}
\end{figure}

\begin{figure}[!t]
\centering
\subfloat[Recall  vs. number of proposals]{
\includegraphics[width=0.5\linewidth]{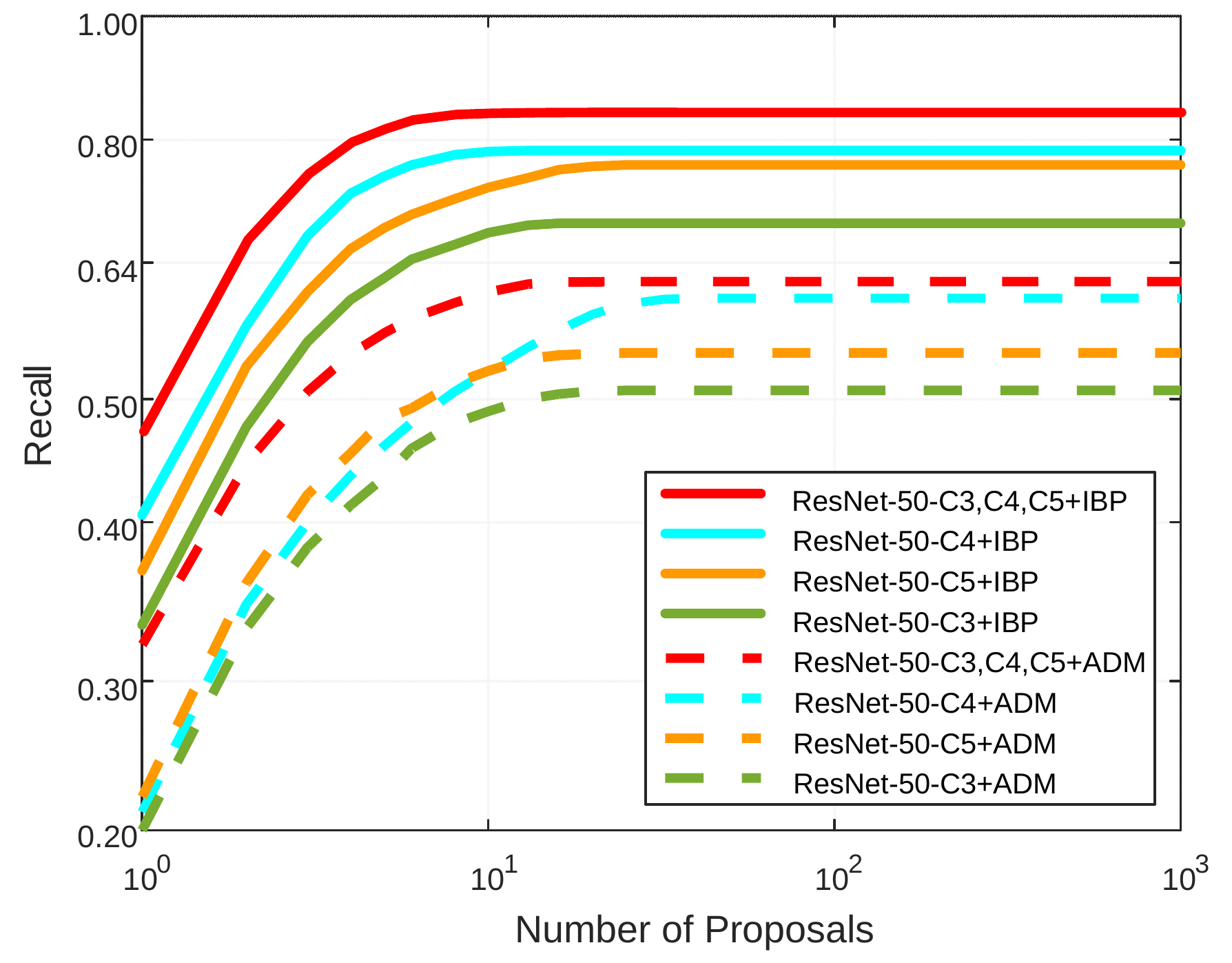}
}
\subfloat[Precision vs. recall]{
\includegraphics[width=0.5\linewidth]{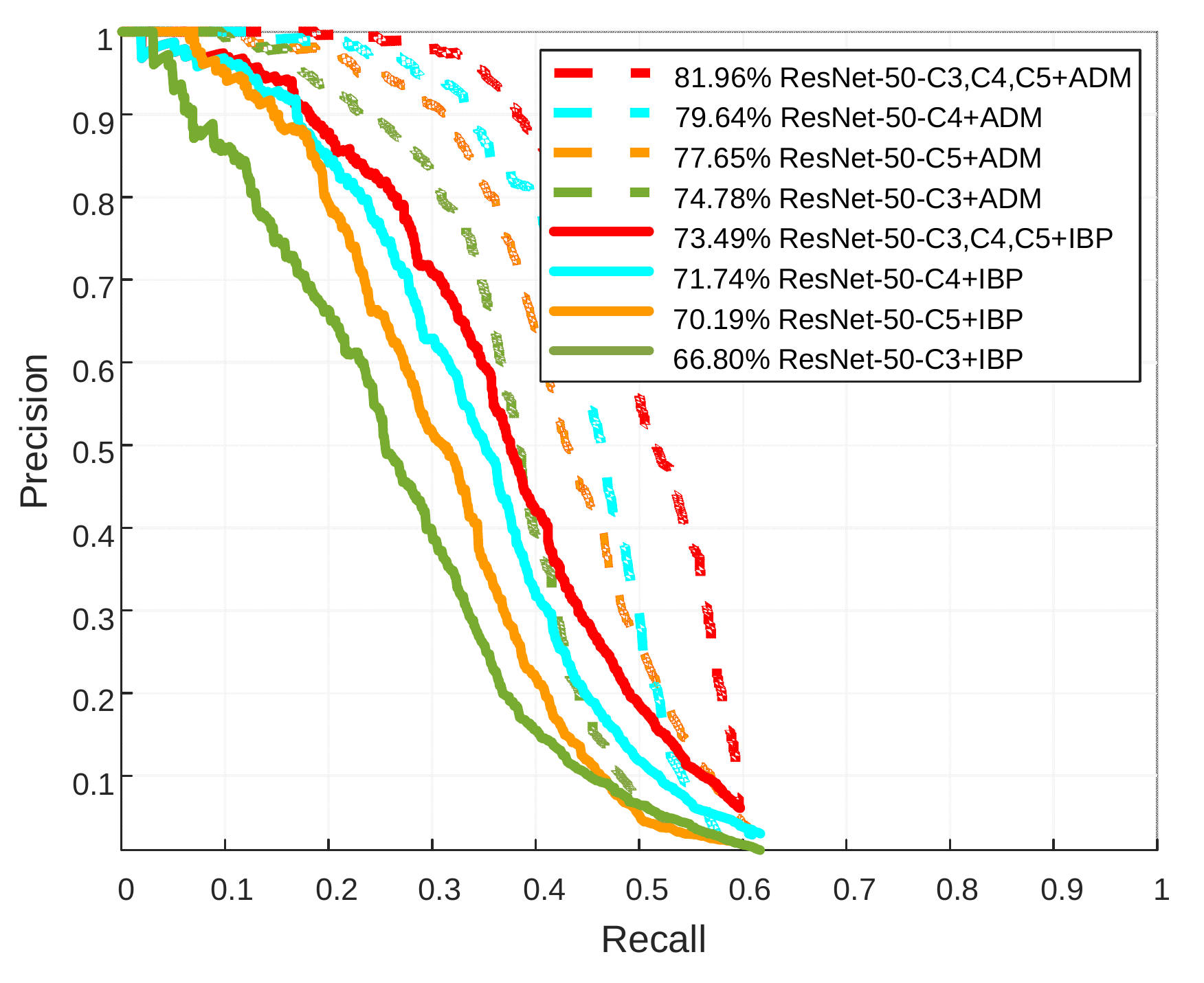}
}
\caption{Recall as a function of the number of pedestrian proposals in Caltech benchmark.
Here ADM refers to the active detection module, IBP refers to producing the initial bbox proposals. See text for details.}
\label{fig_AMD}
\end{figure}

\section{Experiments}
Without loss of generality, our feature representation employed in this paper is based on a ResNet-50 network~\cite{HeEtAl:cvpr16} pre-trained on Imagenet,
where the convolutional layers and max pooling layers of the ResNet are engaged at multiple locations to produce multi-layer feature representations of the input image.

The initial bbox proposal network as illustrated in Fig.~\ref{fig_2} is trained by stochastic gradient descent (SGD) with a learning rate of 0.001 and a momentum of 0.9 for 20k mini-batches, and with weight decay setting to 0.0005.
As evidenced in e.g.~\cite{RenEtAl:nips15,DaiEtAl:nips16} that a larger set of proposals (e.g., 2000) per image has almost no additional benefit,
the number of initial proposals is set to 300 in our approach.

Each mini-batch consists of 128 randomly sampled pedestrian proposals from one randomly selected image, which is composed of 32 positive proposals and 96 negative proposals.
During training and performance evaluation,
a pedestrian bbox prediction is considered to be positive, if its intersection over union with a ground-truth bbox is larger than 0.5, or it has the highest IoU overlap with a ground-truth bbox; It is assigned negative if its IoU ratio is less than 0.3 for any of the ground-truth bboxes.
In our active detector module, each RNN state vector $\vec{s}_t \in \mathbb{R}^{64}$ contains 64 LSTM units of history records. Meanwhile the feature vector obtained from the RoI pooling layer of ResNet-50 delineated by previous bbox prediction is $\vec{o}_t \in \mathbb{R}^{1024}$.
During reinforcement learning training phase, the learning rate is linearly annealed from its initial value (usually between 0.1 and 0.0001) to 0.

For benchmarks, three widely used datasets are considered in our experiments, namely Caltech~\cite{DolEtAl:cvpr09}, ETH~\cite{DolEtAl:bmvc10} and TUD-Brussels~\cite{WojWalSch:cvpr09}.
The Caltech pedestrian dataset consists of approximately 10 hours of 640$\times$480 30Hz video footage taken from a vehicle driving through regular traffic in urban areas,
which finally amounts to about 250,000 frames over around 2,300 unique pedestrians.
The ETH benchmark dataset contains 3 testing video sequences with a resolution of 640$\times$480, and a frame rate of 13FPS.
Finally, the TUD-Brussels pedestrian dataset is recorded from a driving car with a resolution of 640$\times$480 in the inner city of Brussels.
For all the comparison methods considered, their own publicly available results are directly put in use throughout our experiments.

\begin{figure}[!t]
\centering
\subfloat[Near-scale (height$\geq$80 pixels)]{
\includegraphics[width=0.49\linewidth]{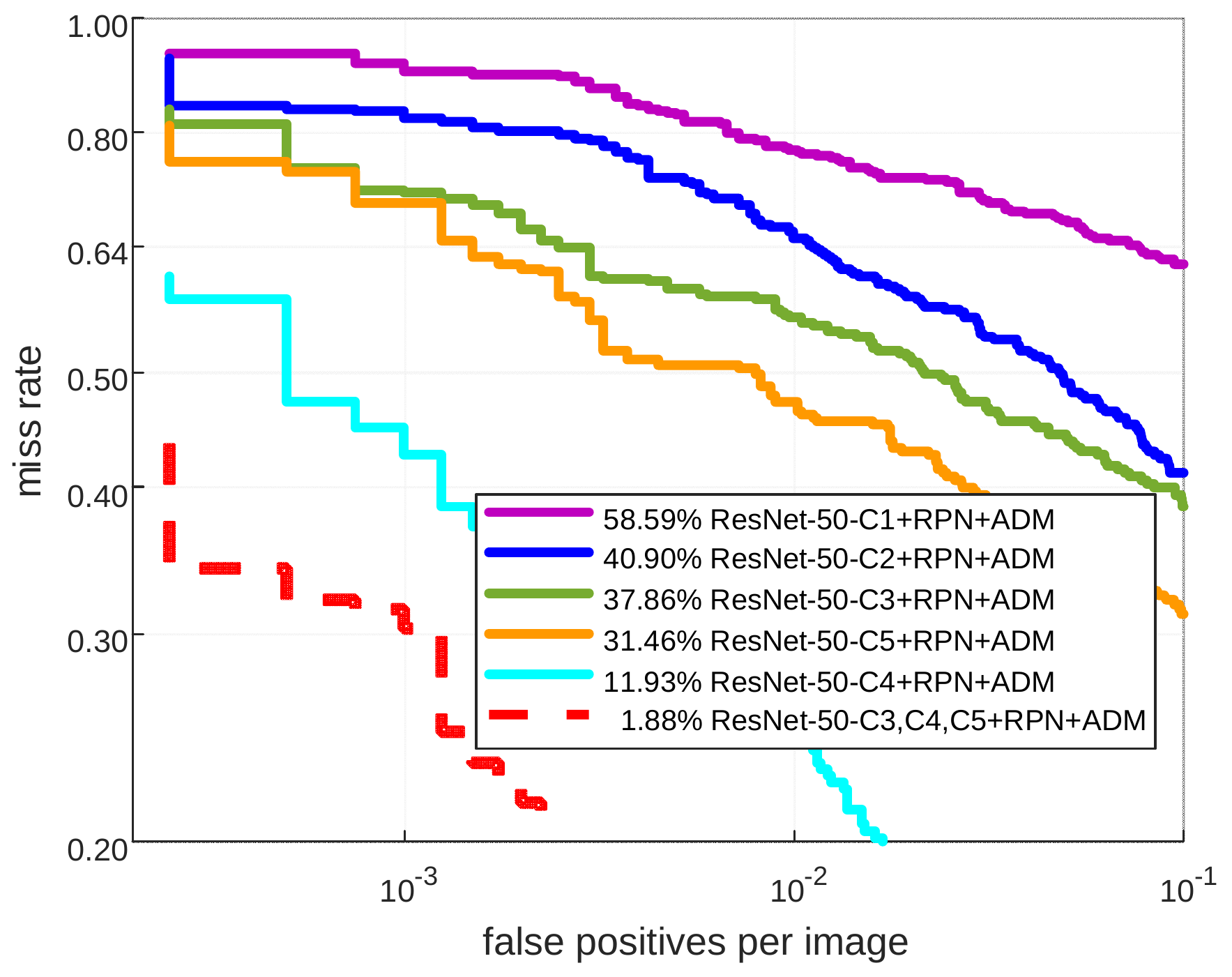}
}
\subfloat[Far-scale (height$<$80 pixels)]{
\includegraphics[width=0.49\linewidth]{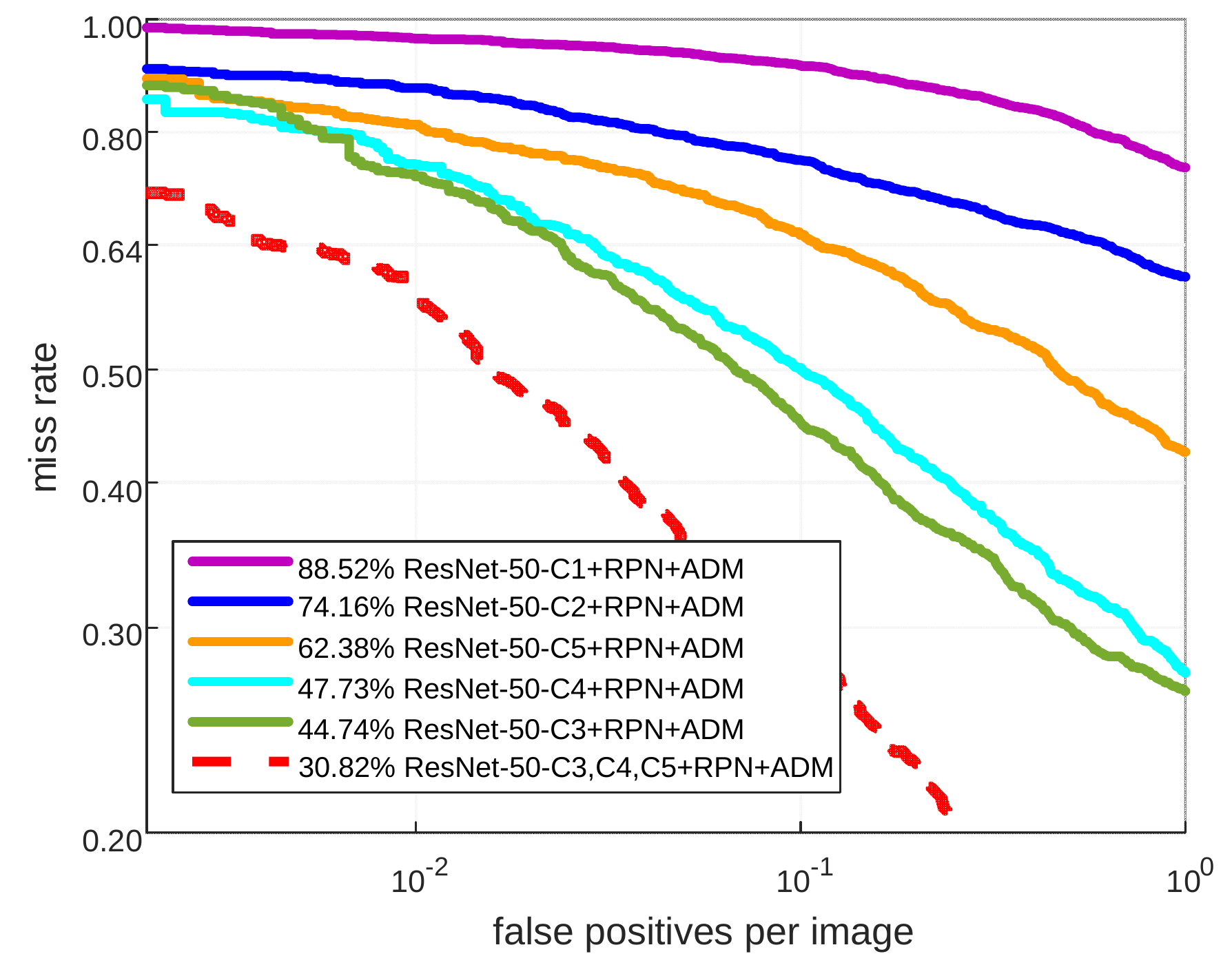}
}
\caption{Performance (in terms of miss rate) of our active detector model when operating in different representation spaces for Caltech benchmark.
Here ADM refers to active detection module, IBP refers to producing the initial bbox proposals.}
\label{fig_9}
\end{figure}

\subsection{Ablation Experiments}
We evaluate the effectiveness of the two major modules in our approach, namely,
the convolutional network for initial proposals and multi-layer representations and the active pedestrian detector.
Without loss of generality, ablation experiments are carried out on the widely used Caltech benchmark.
Three ablation scenarios are examined:
First, we look into the effect of removing the second module.
In other words, our system at this moment is reduced to directly use the initial proposals from multi-layer representations as the final output,
which is accomplished without the follow-up active detection module;
Second, we consider the scenario of having only our second module up and running, i.e. without the initial proposals and multi-layer representation;
Finally, the impact of restricting our active detector to operate on smaller representation spaces is investigated.
Experimental results demonstrate the significance of retaining both modules as essential ingredients of our approach.

\begin{figure*}[!t]
\centering
\subfloat[Near-scale (pedestrian height$\geq$80 pixels)]{
\includegraphics[width=0.33\linewidth]{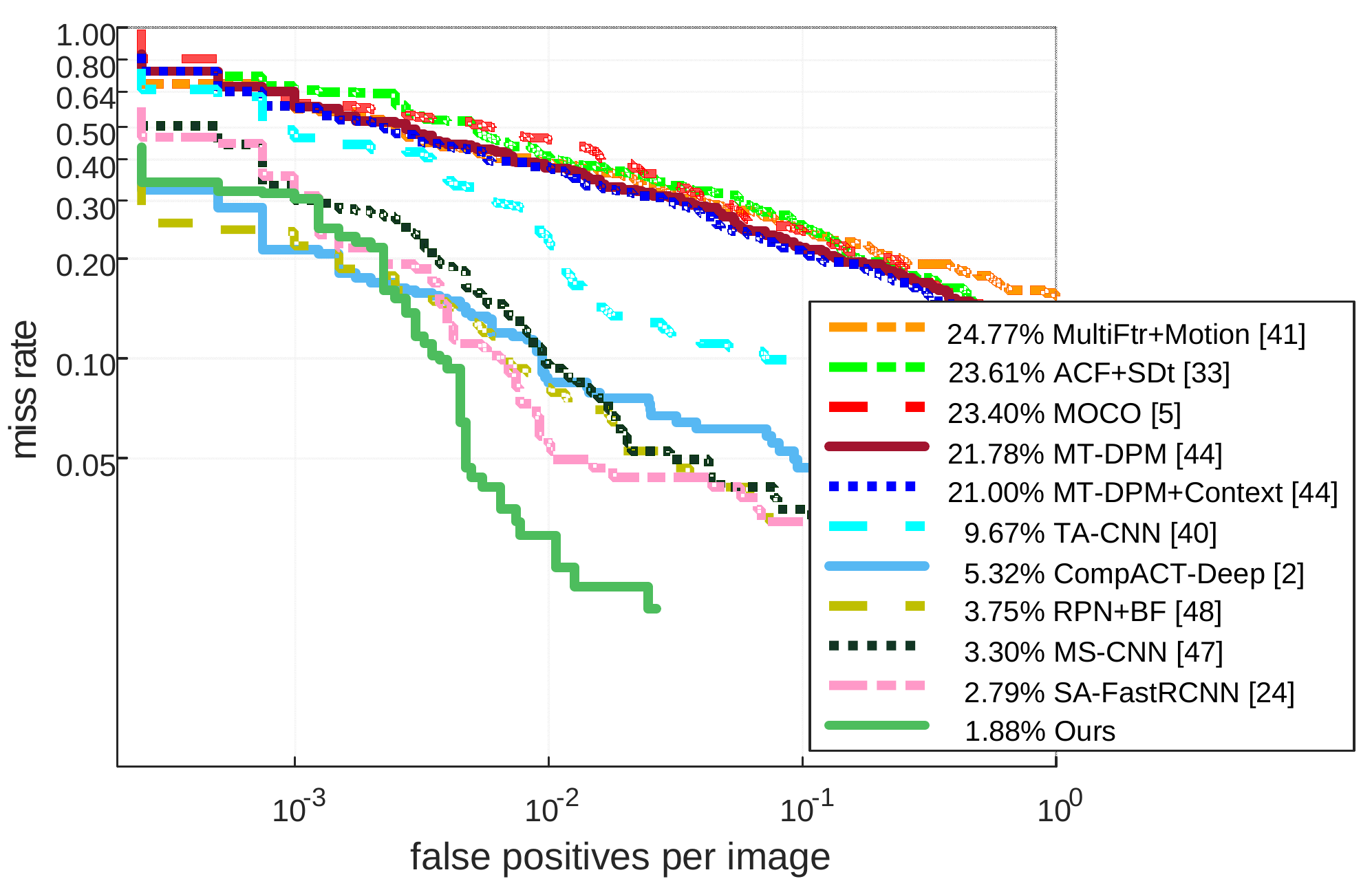}
}
\subfloat[Far-scale (pedestrian height$<$80 pixels)]{
\includegraphics[width=0.33\linewidth]{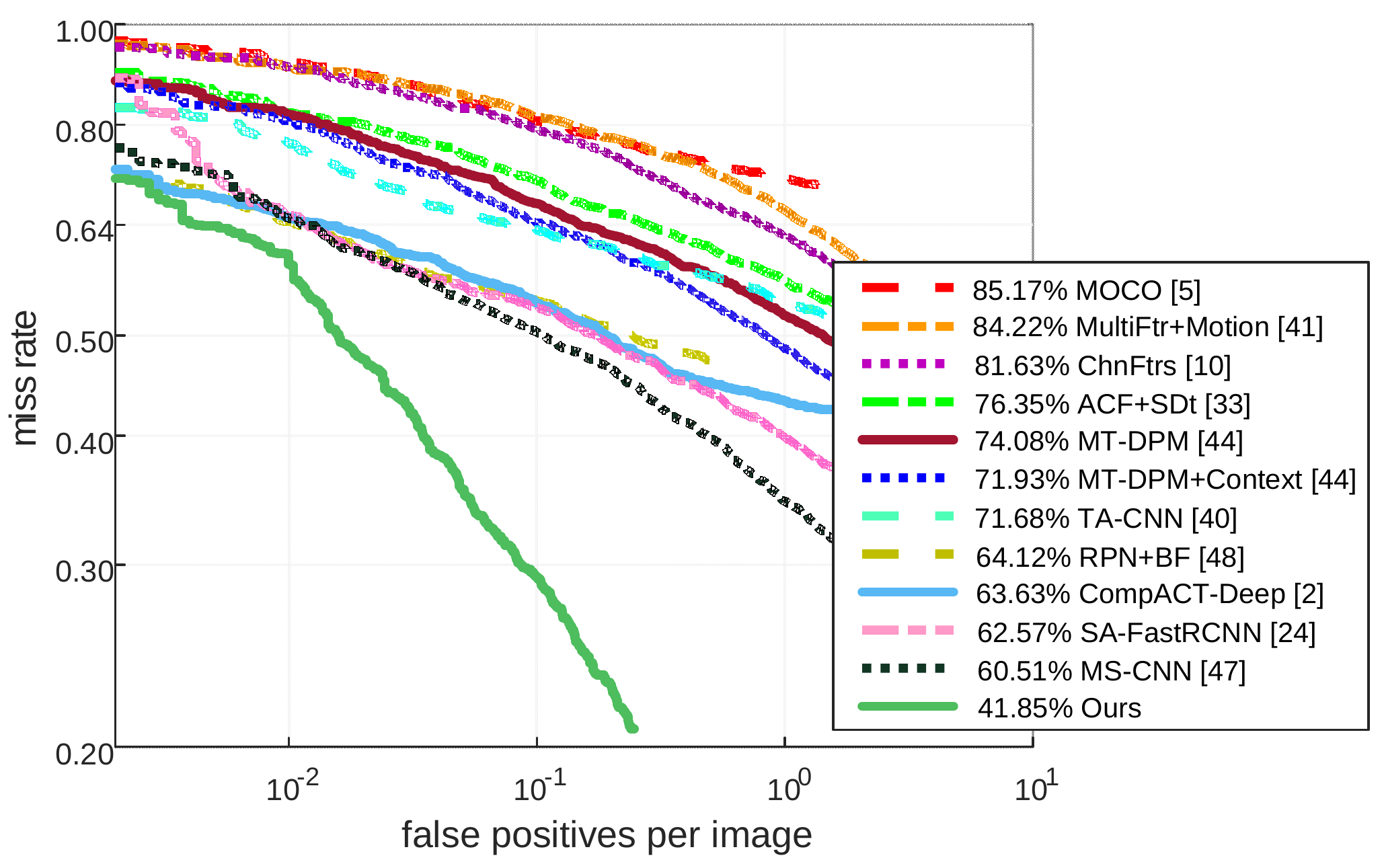}
}
\subfloat[Overall]{
\includegraphics[width=0.33\linewidth]{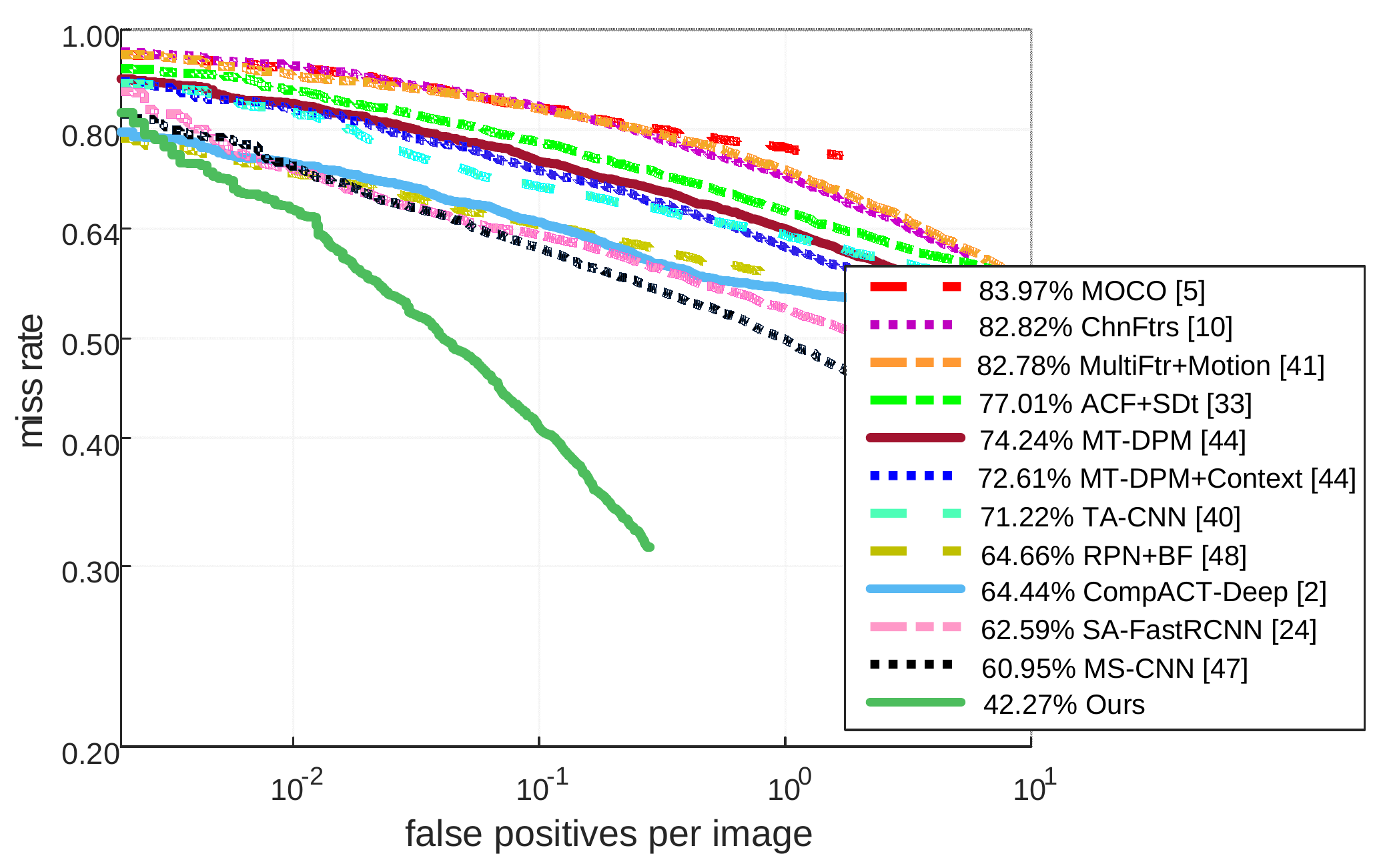}
}
\caption{Quantitative comparison results on the Caltech benchmark.}
\label{fig_13}
\end{figure*}

\begin{figure}[!t]
\centering
\includegraphics[width=0.99\linewidth]{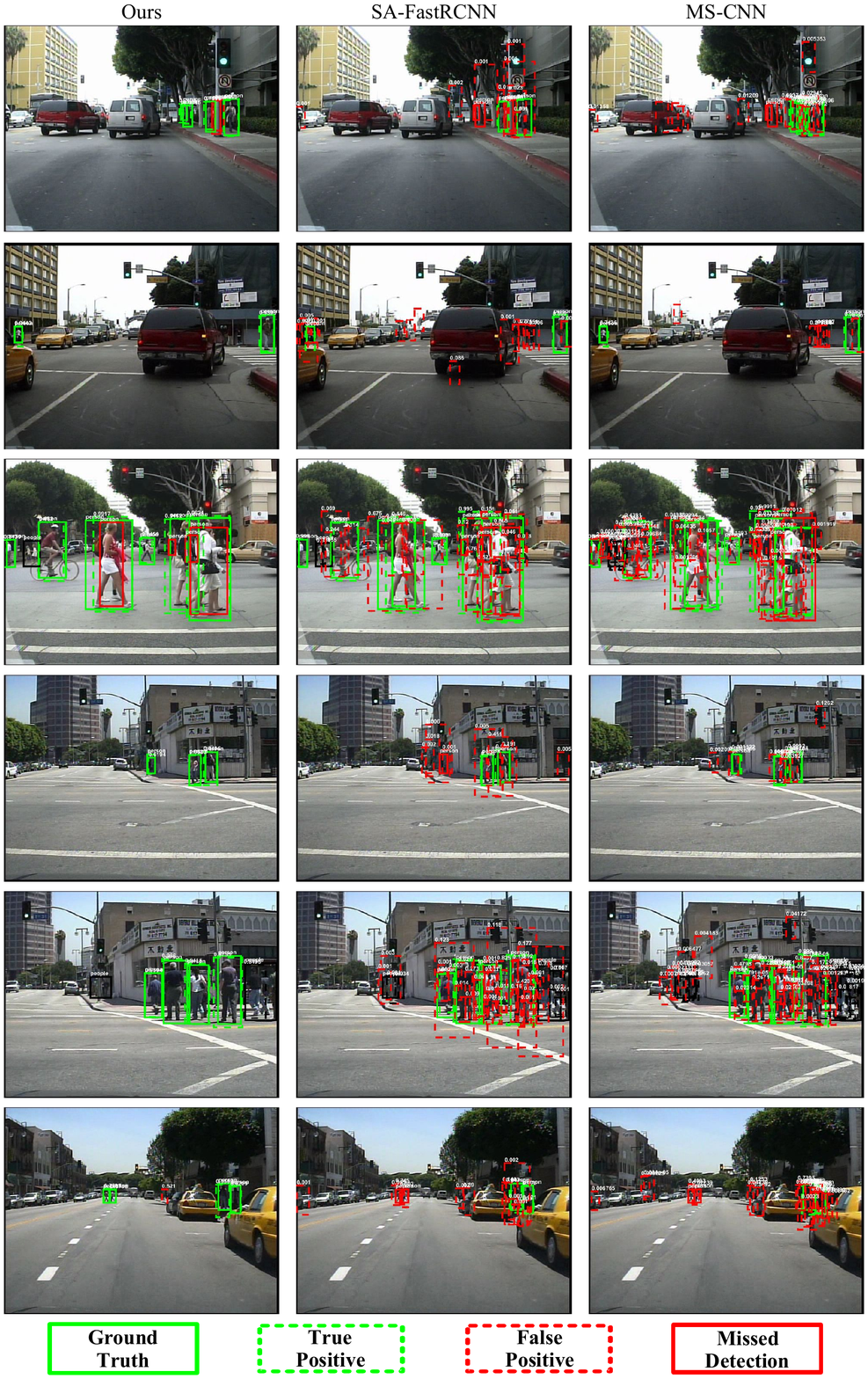}\\
\caption{Visual comparison of our detection results vs. those of the state-of-the-arts on the Caltech benchmark.}
\label{fig_Caltech_CompareRes}
\end{figure}

\subsubsection{Having only the first or the second module}
We start by examining the impact of employing only the first module, i.e. the initial proposals produced from different representation spaces as the final output,
thus without engaging the active detection module.
This is evaluated in terms of the recall rate over IoUs, which primarily relies on the quality of our initial proposals.
Empirically we look at each convolutional layer (from ResNet-50-C1 to ResNet-50-C5) of ResNet~\cite{HeEtAl:cvpr16} for the 300 pedestrian proposals obtained from each image.
As illustrated in Fig.~\ref{fig_7}, overall the first two layers, namely ResNet-50-C1 and ResNet-50-C2, consistently produce poor recall rates across different scales,
which can be explained by the weaker representation capacity of the shallower layers.
On the other hand, higher convolutional layers can encode more of the semantic-level information of the pedestrians,
and such representations are relatively robust to appearance variations.
When working with near-scale pedestrians, higher convolutional layer such as ResNet-50-C4 with a recall rate (by fixing to IoU=0.5) of 95.54\% stands out for delivering better proposals;
Meanwhile for far-scale pedestrians, the performance of higher convolutional layers are much worse, mostly due to their difficulty in localizing small-size objects,
on the other hand, lower layers such as ResNet-50-C3 perform best at a recall rate of 57.91\%.

Nevertheless, the multi-layer representation always delivers best results regardless of the pedestrian instances being at near- or far-scales,
at their respective recall rates of 97.13\% and 68.74\%.
This stresses the importance of adopting a multi-layer representation for the pedestrian detection task.

The impact of engaging only our active detection module (i.e. second module) is also discussed.
Fig.~\ref{fig_AMD}(b) illustrates that overall the partial system with only the second module works noticeably better in comparison with engaging only the first module.
For example, compared to 73.49\% the peak F1 score of engaging only the first module with multi-layer feature representation,
the performance of deploying only the second module with multi-layer feature representation rises to 81.96\%.
On the other hand, the main challenge with the active detection module is to improve its recall rate,
as it usually lags behind the partial system when only the first module is used, as illustrated in Fig.~\ref{fig_AMD}(a).
It is worth noting that Fig.~\ref{fig_AMD}(a) also empirically suggests that the recall rate is oblivious to the introduction of more bbox proposals,
as long as there is a sufficient number of bbox proposals to work with presently.

\subsubsection{Operating our active detector on different representation spaces}
Here the focus is toward inspecting the effect of our active detector operating on different representation spaces, that is, on individual feature layers,
as well as the collective multi-layer representations. Again, without loss of generality we consider here ResNet-50-C1 to ResNet-50-C5 of ResNet~\cite{HeEtAl:cvpr16}.
As displayed in Fig.~\ref{fig_9}, it is clear that the first two layers namely ResNet-50-C1 and ResNet-50-C2 are not the ideal representation spaces for our localization policy,
which is to be expected due to the same set of reasons explained previously.
In terms of performance on single layers, it is once again empirically verified that far-scale instances are best captured by lower layer ResNet-50-C3, with a miss rate 44.74\%;
Similarly for near-scale instances, upper layer ResNet-50-C4 is picked up as the best single layer with a miss rate of 11.93\%.
In comparison, we need to also note down the best results obtained by engaging only the first module as described in the previous section,
which gives miss rates of 40.25\% and 3.15\% respectively on far- and near-scale pedestrian instances.
This may suggest that in our problem context,
detections from multi-layer representation alone perhaps make greater contribution than localization policy acting on single layer representations with associated initial proposals.
Finally, when executing our full approach, the miss rates are substantially reduced to 30.82\% and 1.88\%, respectively.
They demonstrate that both multi-layer representation and localization policy play indispensable roles in our approach.

\begin{figure*}[!t]
\centering
\subfloat[Near-scale (pedestrian height$\geq$80 pixels)]{
\includegraphics[width=0.33\linewidth]{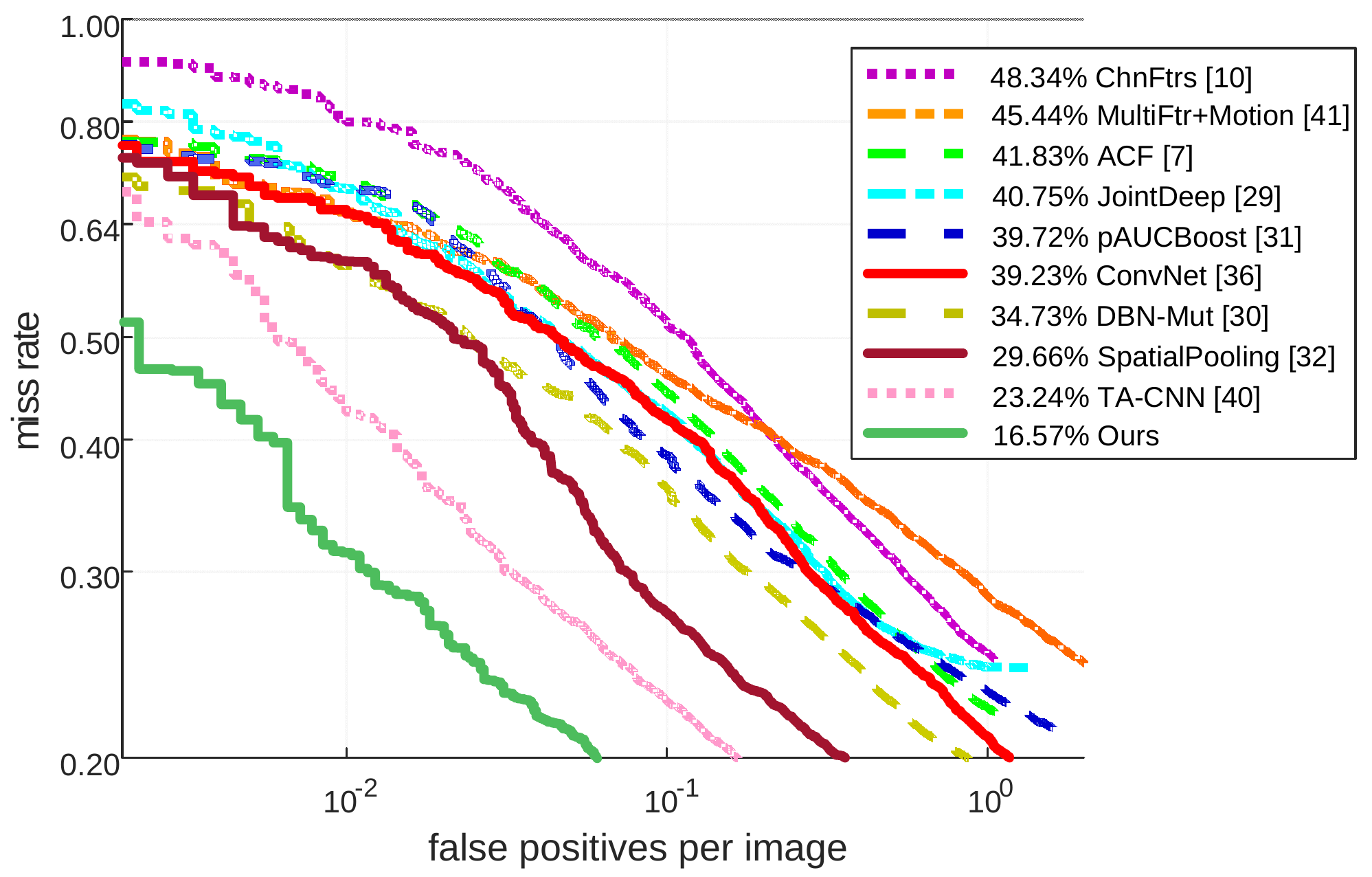}
}
\subfloat[Far-scale (pedestrian height$<$80 pixels)]{
\includegraphics[width=0.33\linewidth]{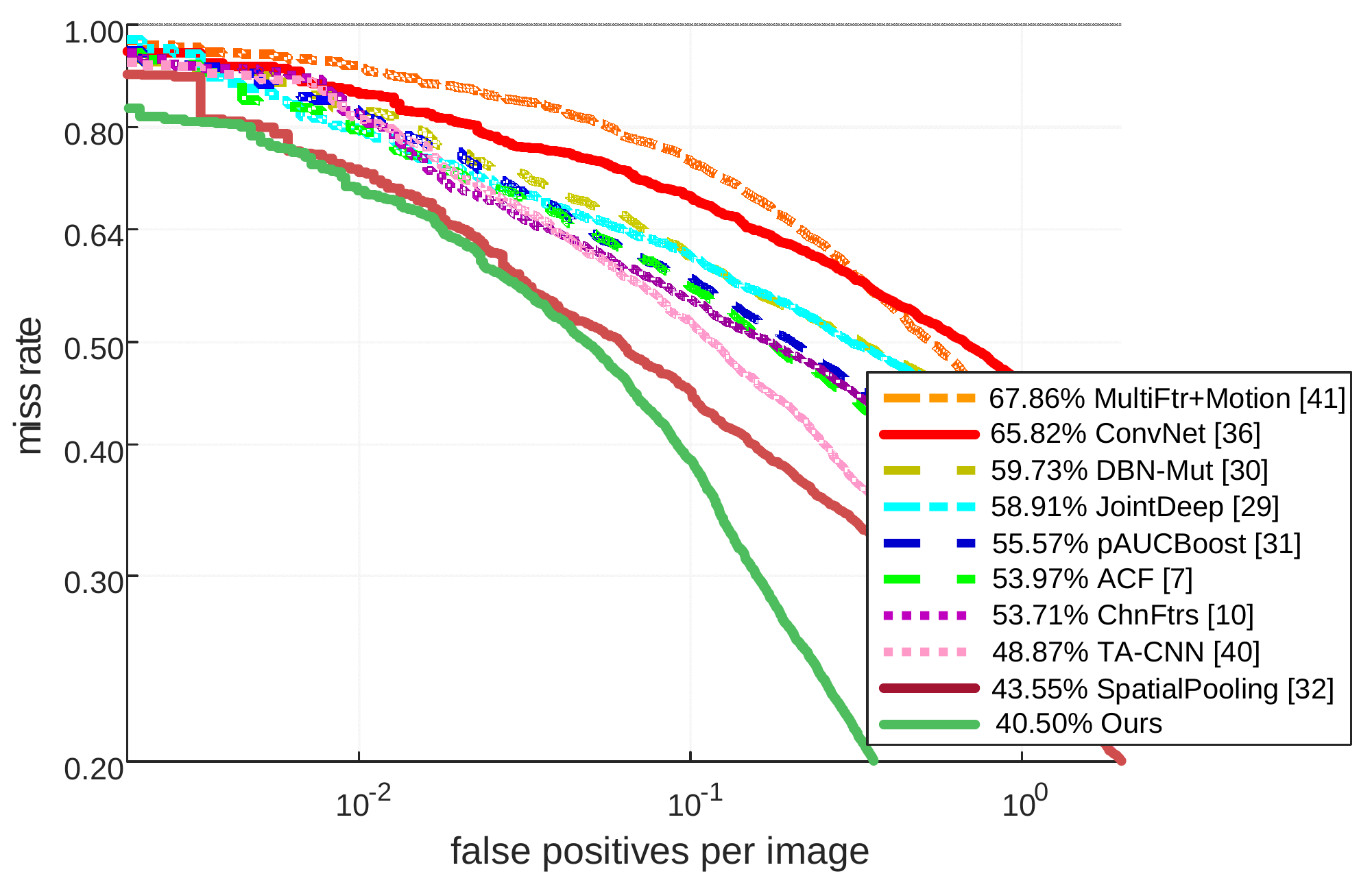}
}
\subfloat[Overall]{
\includegraphics[width=0.33\linewidth]{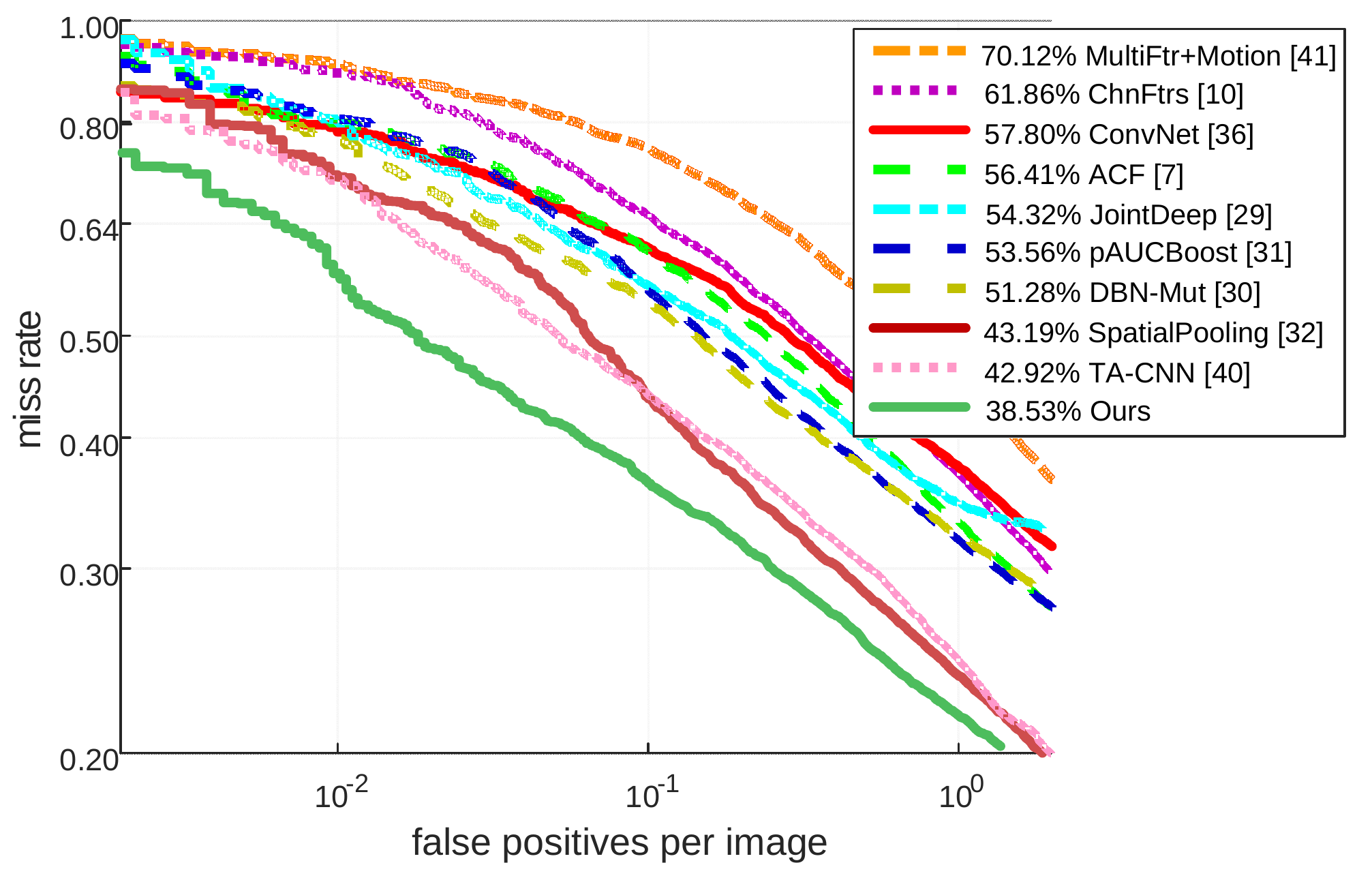}
}
\caption{Quantitative comparison results on the ETH benchmark.}
\label{fig_10}
\end{figure*}

\begin{figure}[!t]
\centering
\includegraphics[width=0.99\linewidth]{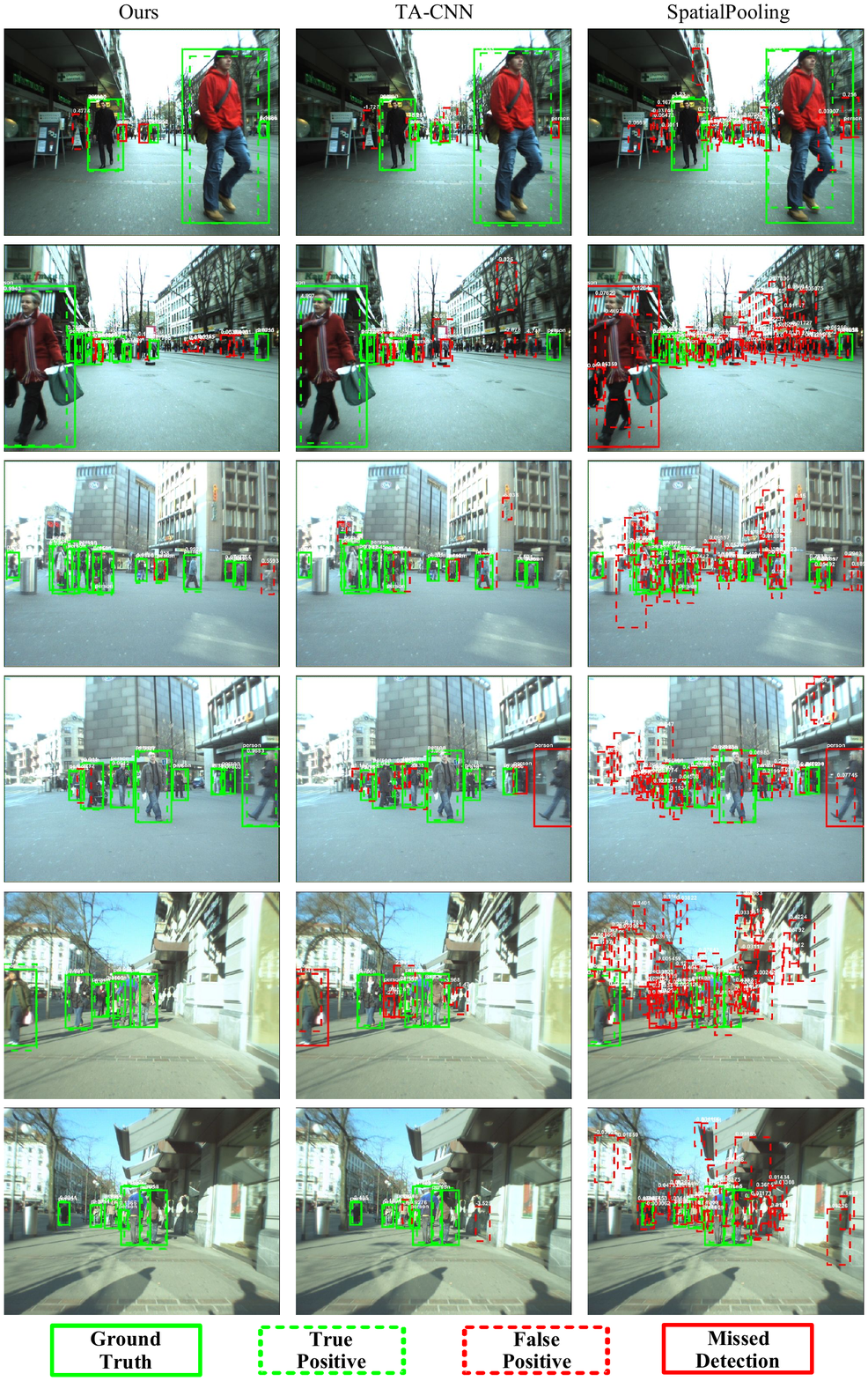}\\
\caption{Visual comparison of our detection results vs. those of the state-of-the-arts on the ETH benchmark.}
\label{fig_ETH_CompareRes}
\end{figure}

\subsection{Comparing with State-of-the-art methods on different benchmarks}
Here we conduct thorough examinations of the proposed approach,
and analyze its performance on widely used benchmarks including Caltech~\cite{DolEtAl:cvpr09}, ETH~\cite{DolEtAl:bmvc10} and TUD-Brussels~\cite{WojWalSch:cvpr09}.
In terms of evaluation metric, we follow the convention and employ the log-average miss rate to summarize the detector performance.
It is obtained by averaging miss rate at FPPI (false positives per-image) rates evenly spaced in log-space within the range of $10^{-3}$ to $10^0$.
In summary, our approach has been demonstrated to consistently outperform the state-of-the-arts over these benchmarks, and perform exceptionally well with far-scale instances.

\subsubsection{Caltech benchmark}
Following the typical protocol adopted by existing literature~\cite{RedEtAl:cvpr16,TiaEtAl:iccv15,YanEtAl:cvpr14,CaiSabVas:iccv15,ZhaBenSch:cvpr15,HosEtAl:cvpr15},
The training phase of our approach is realized at a joint training set consisting of both Caltech and INRIA training images,
and the test phase evaluation is worked out at Caltech testing set.
For comparison, we enlist here a set of ten state-of-the-art methods, including MultiFtr+Motion~\cite{WalEtAl:cvpr10}, ACF+SDt~\cite{ParEtAl:cvpr13},
MOCO~\cite{CheEtAl:cvpr13}, MT-DPM \& MT-DPM+Context~\cite{YanEtAl:cvpr14}, TA-CNN~\cite{TiaEtAl:cvpr15}, CompACT-Deep~\cite{CaiSabVas:iccv15}, RPN+BF~\cite{ZhaEtAl:eccv16}, MS-CNN~\cite{CaiEtAl:eccv16}, and SA-FastRCNN~\cite{LiEtAl:cvpr15}.
Comparison results are evaluated in terms of the log-average miss rate for pedestrian instances of three scenarios:
(a) near-scale, i.e. no less than 80 pixels in height, (b) far-scale, i.e. shorter than 80 pixels, and (c) all, which is a combination of both.

Fig.~\ref{fig_13}(a) displays the quantitative results of near-scale (a).
Our approach outperforms all comparison methods and achieves the lowest log-average miss rate of 1.88\%,
which clearly exceeds the two best existing results: 2.79\% of SA-FastRCNN~\cite{LiEtAl:cvpr15}, and 3.30\% of MS-CNN~\cite{CaiEtAl:eccv16}.
Further, for far-scale (b), our approach achieves the lowest miss rate of 41.85\%, which amounts to substantial better performance than the top two existing results,
namely 62.57\% of SA-FastRCNN and 60.51\% of MS-CNN, as exhibited in Fig.~\ref{fig_13}(b).
Fig.~\ref{fig_13}(c) presents the overall performance, where ours again significantly outperform the rest in a trend similar to that of (b),
which is not surprising as the amount of far-scale instances dominates the overall pedestrian population of Caltech benchmark.
Fig.~\ref{fig_Caltech_CompareRes} presents exemplar detection results of our approach on Caltech test images.
Note that most pedestrians including in particular far-scale instances can now be detected correctly by our approach.
It also provides visual comparisons,
where evidently the state-of-the-art methods such as SA-FastRCNN~\cite{LiEtAl:cvpr15} and MS-CNN~\cite{CaiEtAl:eccv16} produce more false alarms as well as more missings.

\begin{figure*}[!t]
\centering
\subfloat[Near-scale (pedestrian height$\geq$80 pixels)]{
\includegraphics[width=0.33\linewidth]{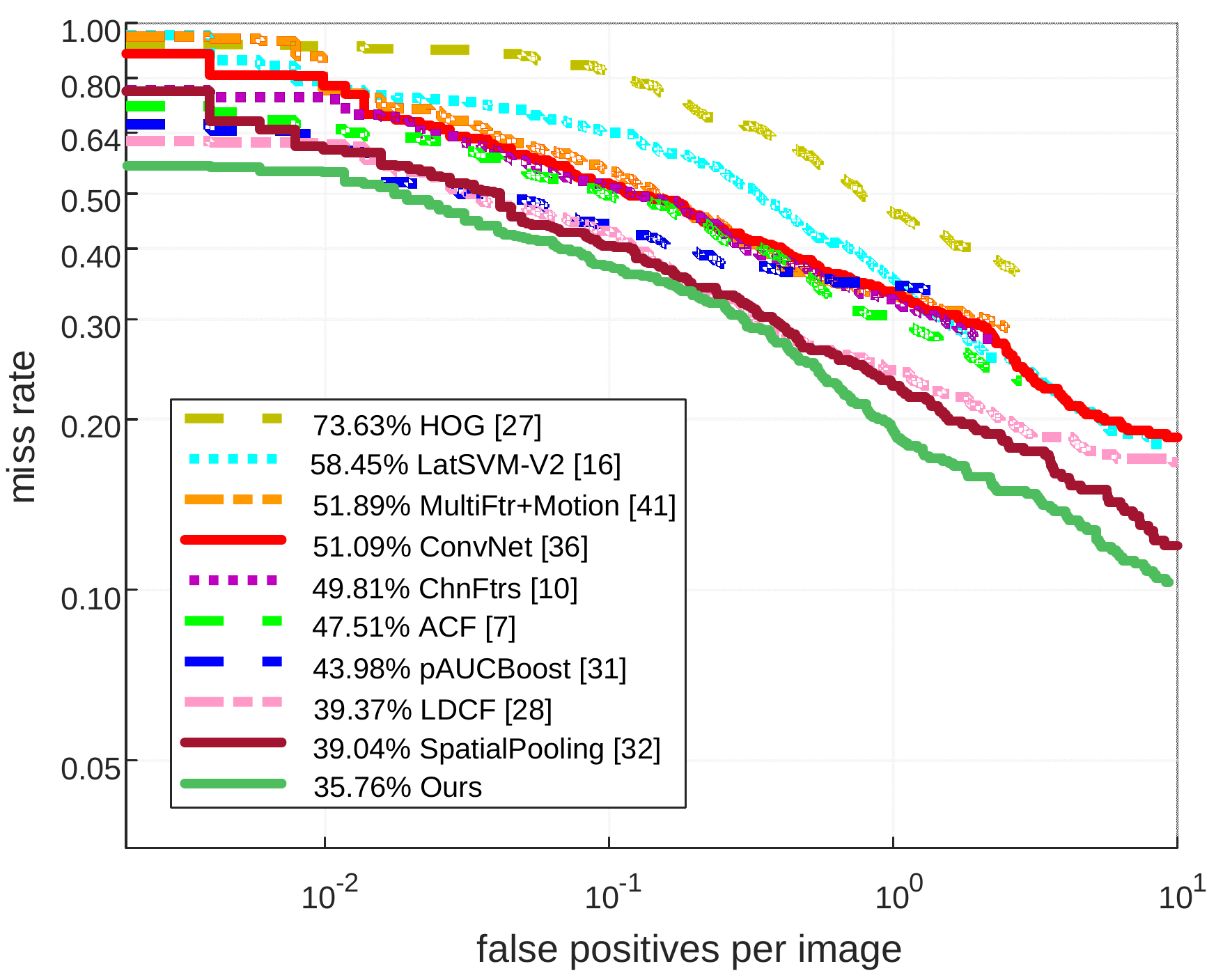}
}
\subfloat[Far-scale (pedestrian height$<$80 pixels)]{
\includegraphics[width=0.33\linewidth]{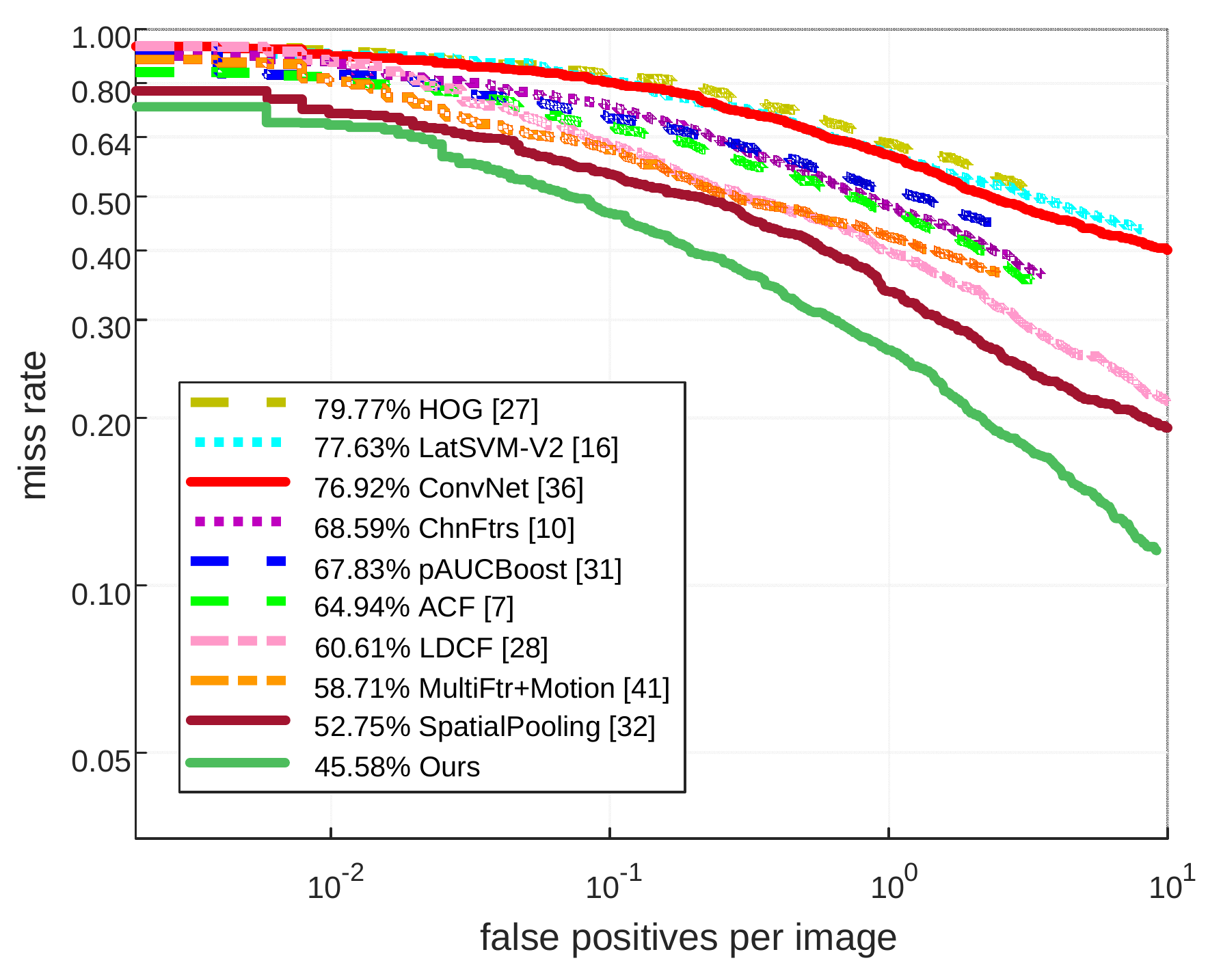}
}
\subfloat[Overall]{
\includegraphics[width=0.33\linewidth]{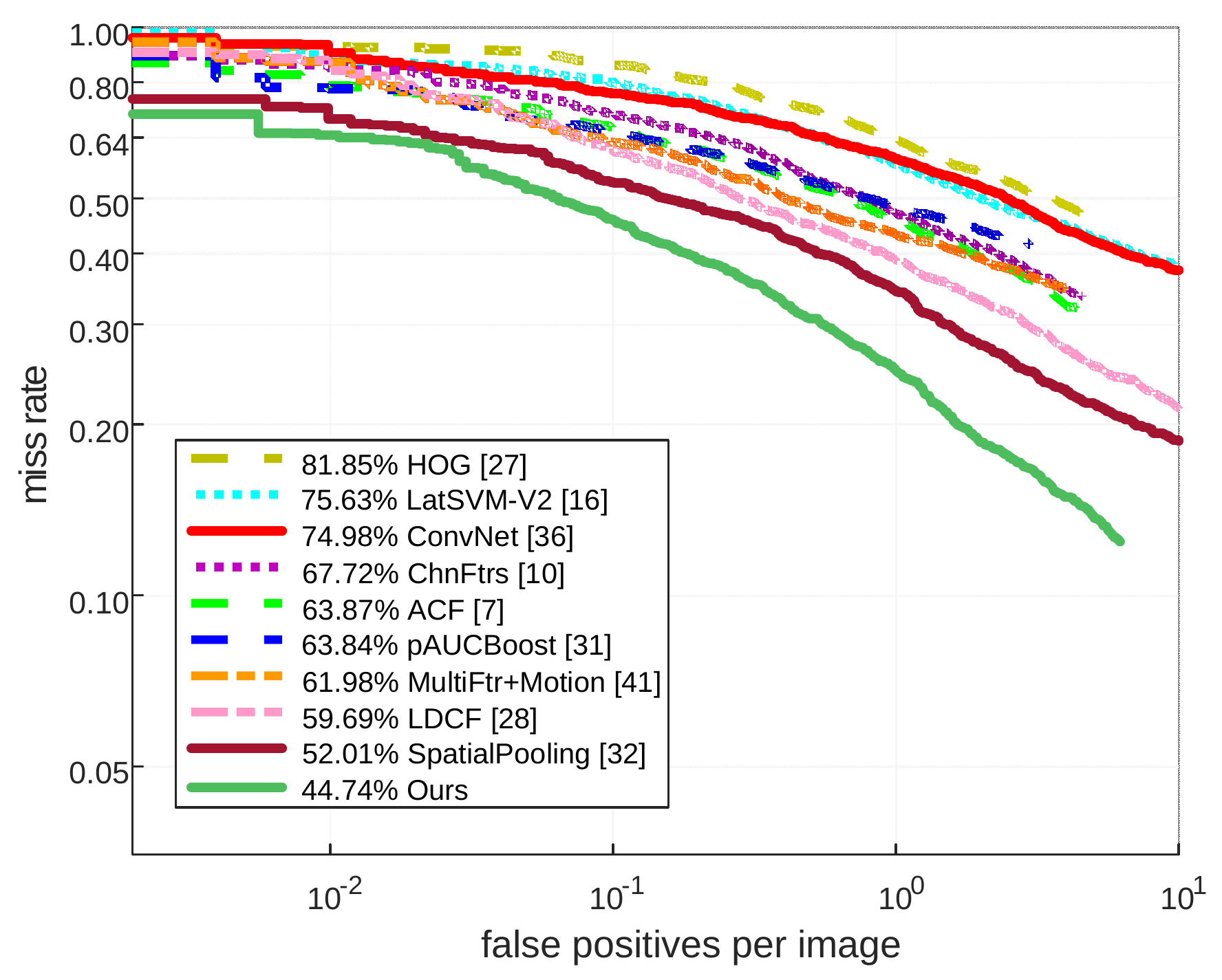}
}
\caption{Quantitative comparison results on the TUD-Brussels benchmark.}
\label{fig_TUD}
\end{figure*}

\begin{figure}[!t]
\centering
\includegraphics[width=0.99\linewidth]{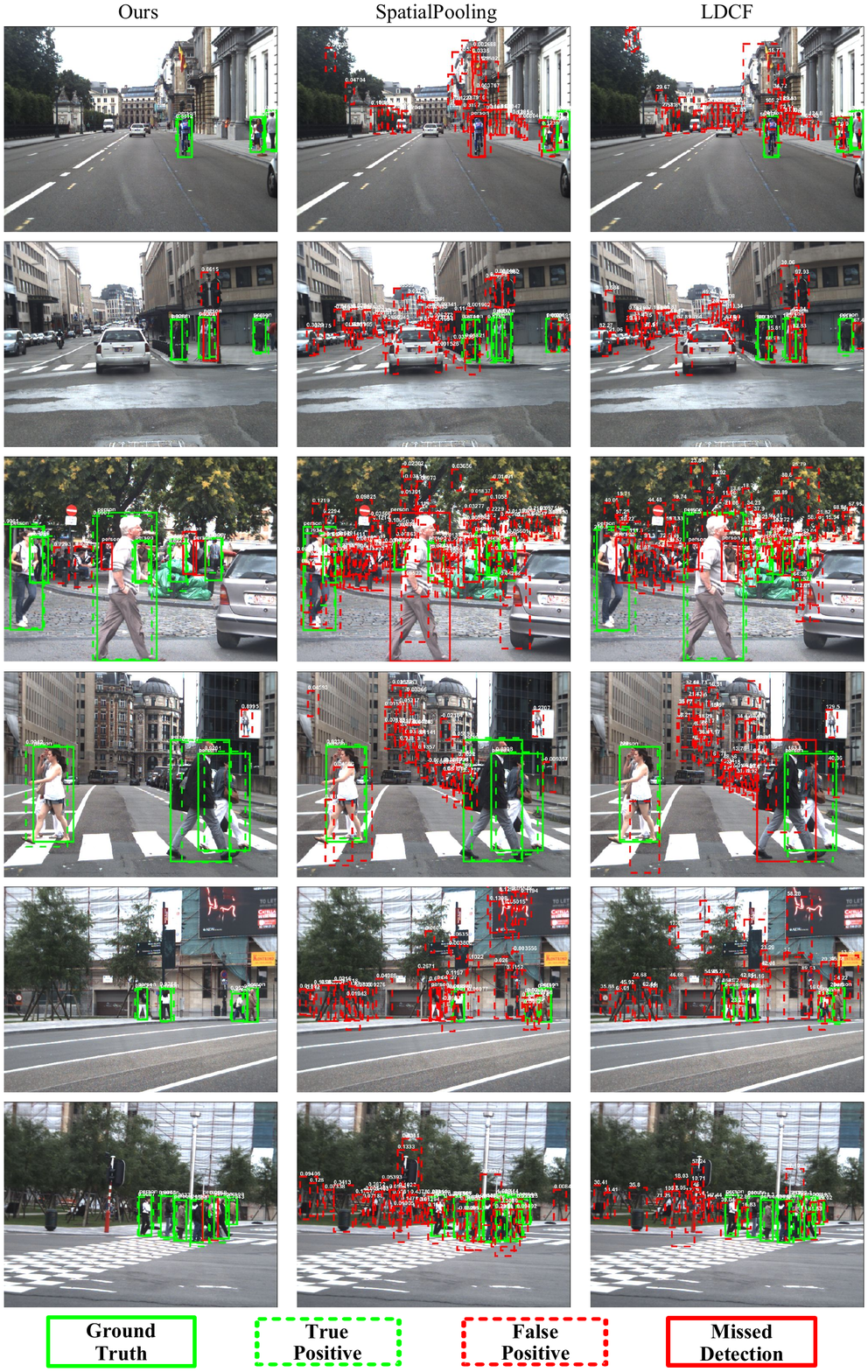}\\
\caption{Visual comparison of our detection results vs. those of the state-of-the-arts on the TUD-Brussels benchmark.}
\label{fig_TUD_CompareRes}
\end{figure}

\subsubsection{ETH benchmark}
Similarly, we follow the convention adopted by most state-of-the-art studies such as~\cite{ZhaEtAl:eccv16,RedEtAl:cvpr16,CaiEtAl:eccv16,Gir:iccv15},
to train our system on INRIA training dataset, which is then deployed on the ETH testset.
Comparison methods include nine state-of-the-art methods, which are ChnFtrs~\cite{DolEtAl:bmvc09}, MultiFtr+Motion~\cite{WalEtAl:cvpr10}, ACF~\cite{DolEtAl:tpami14},
JointDeep~\cite{OuyWan:iccv13}, pAUCBoost~\cite{PaiSheHen:iccv13}, ConvNet~\cite{SerEtAl:cvpr13}, DBN-Mut~\cite{OuyZenWan:cvpr13}, SpatialPooling~\cite{PaiSheHen:eccv14},
and TA-CNN~\cite{TiaEtAl:cvpr15}.
Comparison results are evaluated on the same three pedestrian scenarios stated previously.

As displayed in Fig.~\ref{fig_10}(a), for near-scale pedestrians, the top two best methods, SpatialPooling and TA-CNN, have respective miss rate of 29.66\% and 23.24\%,
which is further reduced to as low as 17.63\% by our approach;
Similar pattern can also be found in Fig.~\ref{fig_10}(b) for the much harder scenario of far-scale instances,
where our approach (40.50\%) outperforms the top two best methods, SpatialPooling (43.55\%) and TA-CNN (48.87\%), by a noticeable margin of over 3\%.
Here the relative small gain is attributed to the inherent challenge of this dataset where many far-scale pedestrians are also severely occluded.
Fig.~\ref{fig_10}(c) reveals the overall performance,
where our approach delivers the least missing rate of 38.53\%, versus 43.19\% of SpatialPooling and 42.92\% of TA-CNN, the two top performers of ETH benchmark.
Exemplar visual results of our approach on ETH dataset is presented in Fig.~\ref{fig_ETH_CompareRes}.
Again most pedestrian instances across distinct scales are correctly detected.
On the other hand, there still exists a relative small fraction of mistakes which usually are these extremely far-scale instances.
Similar to the Caltech case, visual comparisons are also provided in Fig.\ref{fig_ETH_CompareRes},
where the two top performers, SpatialPooling and TA-CNN, again produce significantly more false alarms as well as missing instances than those of our approach.

\subsubsection{TUD-Brussels benchmark}
Now we study the performance on TUD-Brussels benchmark, where the typical protocol of system training on the INRIA training set is applied.
Comparison methods include nine state-of-the-art methods, which are HOG~\cite{DalTri:cvpr05}, LatSVM-V2~\cite{FelEtAl:tpami10}, MultiFtr+Motion~\cite{WalEtAl:cvpr10},
ConvNet~\cite{SerEtAl:cvpr13}, ChnFtrs~\cite{DolEtAl:bmvc09}, ACF~\cite{DolEtAl:tpami14}, pAUCBoost~\cite{PaiSheHen:iccv13}, LDCF~\cite{NamDolHan:nips14},
and SpatialPooling~\cite{PaiSheHen:eccv14}.
Comparison results are evaluated on the same three pedestrian scenarios stated previously.

Similar trend to what we have observed for the other two benchmarks also occurs here:
for near-scale we have evidenced clear though relatively small improvement (over 3\%) over the state-of-the-arts,
our approach achieves 35.76\% miss rate where the top competing method is SpatialPooling (39.04\%);
Meanwhile for far-scale instances, the gap between ours (45.58\%) and that of the top performer -- SpatialPooling (52.75\%) is much wider (over 7\%);
As majority of the pedestrians are far-scale ones, the overall performance gap is also on the large side, 44.75\% of ours vs. 52.01\% of the best performer (SpatialPooling),
which amounts to over 7\%.
Note SpatialPooling is also among the top performers for ETH benchmark.
In addition, Fig.~\ref{fig_TUD_CompareRes} displays several visual examples of our approach, as well as side by side visual comparisons. 
Similarly,
the two top performers here, SpatialPooling and LDCF, again produce considerably more false alarms and missing instances than those of our approach.

\section{Conclusion and Outlook}
An effective approach is presented for detecting pedestrian instances of different scales in still images with complex street scenes.
This is realized by an active detection model that bases on a set of initial bounding box proposals,
executes sequences of coordinate transformation actions across multi-layer feature representations to deliver accurate prediction of pedestrian locations.

Empirical examinations over various widely used benchmarks have demonstrated the superior performance of our approach.
For future work, we plan to further address the challenged of detecting heavily occluded pedestrian instances,
as well as investigating the ability of our approach in tackling more generic object detection scenarios.

\section*{Acknowledgment}
This work was partially supported by the National Key Research and Development Program of China (Grant No.2016YFC0801003), the National Natural Science Foundation of China (No. 61370121), and the A*STAR JCO grants 15302FG149 and 1431AFG120. We thank Nastaran Okati for helping with the presentation. 

\bibliographystyle{plain}
\bibliography{objDet_TIP17}

\end{document}